\documentclass{article}

    \PassOptionsToPackage{numbers, compress}{natbib}

\usepackage[final]{neurips_2025}

\usepackage[utf8]{inputenc} 
\usepackage[T1]{fontenc}    
\usepackage{url}            
\usepackage{booktabs}       
\usepackage{amsfonts}       
\usepackage{nicefrac}       
\usepackage{microtype}      
\usepackage{xcolor}         

\usepackage[utf8]{inputenc} 
\usepackage[T1]{fontenc}    
\usepackage{url}            
\usepackage{microtype}      
\usepackage{xcolor}         
\usepackage{xspace}
\usepackage{graphicx,float,subcaption}
\usepackage{amsmath,amssymb,amsfonts,dsfont,pifont,bm,bbm,mathrsfs,mathtools,nicefrac}
\usepackage{algorithm,algpseudocode,listings}
\usepackage{booktabs,multirow,adjustbox,diagbox,threeparttable}
\usepackage{makecell}
\usepackage{wrapfig}
\usepackage{colortbl}
\definecolor{citeblue}{RGB}{48,111,186}
\usepackage{xcolor}
\definecolor{myColor}{RGB}{48,111,186}
\usepackage[pagebackref=false,breaklinks=true,colorlinks=true,citecolor=citeblue,bookmarks=false]{hyperref}
\usepackage{cleveref}  
\usepackage[hypcap=false]{caption}
\usepackage{enumitem}
\usepackage{anyfontsize}
\usepackage{placeins}

\definecolor{mygray}{rgb}{0.9, 0.9, 0.9}

\newcommand{\workname}{ContextAgent}
\newcommand{\benchmark}{ContextAgentBench}
\newcommand{\benchmarklite}{ContextAgentBench-Lite}

\title{\workname: Context-Aware Proactive LLM Agents with Open-World Sensory Perceptions}

{
\setcounter{footnote}{1}%

\author{
    Bufang Yang$^{1}$\thanks{Equal Contribution. $^\ddag$Corresponding Author.} \
    ~~Lilin Xu$^{2}$$^\dag$ \
    Liekang Zeng$^{1}$ \
    Kaiwei Liu$^{1}$ \
    Siyang Jiang$^{1}$ 
     Wenrui Lu$^{1}$ \\
    \textbf{
    Hongkai Chen$^{1}$ \
    Xiaofan Jiang$^{2}$ \
   Guoliang Xing$^{1}$ \
   Zhenyu Yan$^{1\ddag}$
    }
    \\[3pt]
    $^1$The Chinese University of Hong Kong \ $^2$Columbia University 
    \\[3pt]
    \small\texttt{\{bfyang,lkzeng,lk022,syjiang,wrlu,hkchen,glxing,zyyan\}@ie.cuhk.edu.hk},\\
    \small\texttt{lx2331@columbia.edu}, \texttt{jiang@ee.columbia.edu}\\
}
}

\begin{document}

\maketitle

\begin{abstract}

Recent advances in Large Language Models (LLMs) have propelled intelligent agents from reactive responses to proactive support. 
While promising, existing proactive agents either rely exclusively on observations from enclosed environments (e.g., desktop UIs) with direct LLM inference or employ rule-based proactive notifications, leading to suboptimal user intent understanding and limited functionality for proactive service.
In this paper, we introduce \workname, the first context-aware proactive agent that incorporates extensive sensory contexts surrounding humans to enhance the proactivity of LLM agents.
\workname~first extracts multi-dimensional contexts from massive sensory perceptions on wearables (e.g., video and audio) to understand user intentions. 
\workname~then leverages the sensory contexts and personas from historical data to predict the necessity for proactive services. 
When proactive assistance is needed, \workname~further automatically calls the necessary tools to assist users unobtrusively.
To evaluate this new task, we curate \benchmark, the first benchmark for evaluating context-aware proactive LLM agents, covering 1,000 samples across nine daily scenarios and twenty tools.
Experiments on \benchmark~show that \workname~outperforms baselines by achieving up to 8.5\% and 6.0\% higher accuracy in proactive predictions and tool calling, respectively.
We hope our research can inspire the development of more advanced, human-centric, proactive AI assistants.
The code and dataset are publicly available at \href{https://github.com/openaiotlab/ContextAgent}{https://github.com/openaiotlab/ContextAgent}.

\end{abstract}

\section{Introduction}
\label{sec:introduction}
Large Language Model (LLM) agents are revolutionizing our daily life~\cite{guo2025deepseek}, assisting users with complex tasks such as automated web navigation~\cite{deng2023mind2web,zheng2024gpt,cheng2024seeclick}, software engineering~\cite{yang2024swe,zhang2024codeagent,wei2025swe}, and healthcare services~\cite{abbasian2023conversational,yang2024drhouse,merrill2024transforming}.
While LLM agents are receiving growing attention and adoption, most of them still function in a \textit{reactive paradigm}: They can initiate tasks only upon explicit user instructions and yet lack the autonomy to perceive environments and offer proactive support for users.

To further reduce reliance on instructions and alleviate human cognitive workload, proactive agents emerge, which are capable of initiating tasks without explicit user queries~\cite{zhao2025codinggenie,lu2024proactive,liu2024proactive,
zhang2024ask,zhang2024proagent}.
For example, research on proactive agents have explored coding assistance~\cite{zhao2025codinggenie,lu2024proactive}, conversation participation~\cite{liu2024proactive,yang2025socialmind}, re-asking strategies to reduce ambiguity in user instructions~\cite{zhang2024ask}, and multi-agent cooperation scenarios~\cite{zhang2024proagent,wu2025proai}.
However, their limited ability in open-world perceptions and restricted functionality for proactive service hinders their potential as personal companions.

\textbf{Environmental Perception.}
When explicit user instructions are absent, environment perception is crucial for proactive LLM agents. 
Recent studies ~\cite{lu2024proactive,zhao2025codinggenie} proposed proactive agents for programming assistance, while they require access to specific inputs such as computer screenshots or keyboard inputs.
We argue that an ideal proactive agent should be able to perceive open-world environments in the user’s daily life, utilizing wearable devices such as smart glasses and earphones.
By sharing the same perception as the user, the agent can understand the user's intention and provide services automatically.
Besides, the hands-free nature of these ubiquitous wearable devices aligns well with the mission of proactive agents, freeing both the user’s hands and mind from additional workload.

\begin{wrapfigure}{r}{0.6\textwidth}
\vspace*{-5mm}
\begin{center}
\includegraphics[width=\linewidth]{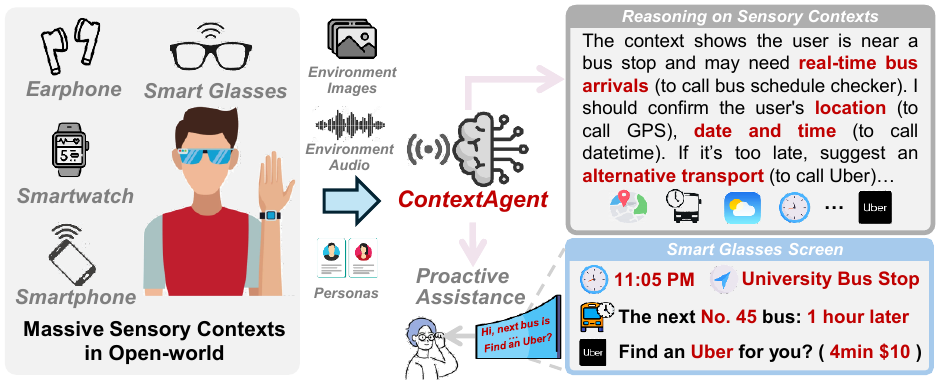}
\end{center}
\vspace{-10pt}
\caption{
\workname~is a proactive AI assistant free of user explicit instructions. 
\workname~can continuously perceive environmental contexts (e.g., image and audio) to detect the necessity of proactive services, and provide tool-augmented assistance based on LLM reasoning.}
\label{fig:teaser_ours}
\vspace{-5pt}
\end{wrapfigure}

\textbf{Functionality for Proactive Services.}
Current personal assistants can deliver proactive notifications via wearables, yet remain limited by static, rule-based pipelines (e.g., alerts when rapid falling is detected~\cite{fall_detection_apple_watch}).
Recent studies~\cite{lu2024proactive,zhao2025codinggenie,yang2025socialmind,zhang2024ask} propose to build proactive agents with LLMs.
However, these agents only provide direct answers during user interactions, without leveraging external tools, and remain limited to enclosed environments (e.g., desktop and keyboard inputs~\cite{lu2024proactive,zhao2025codinggenie}).
Therefore, there remains a research gap in developing a context-aware proactive LLM agent that can exploit extensive sensory contexts to comprehensively understand user intentions, predict the necessity of proactive services, and automatically integrate external tools to deliver unobtrusive services as a personal companion.

\begin{figure}[htb]
    \centering
\includegraphics[width=\linewidth]{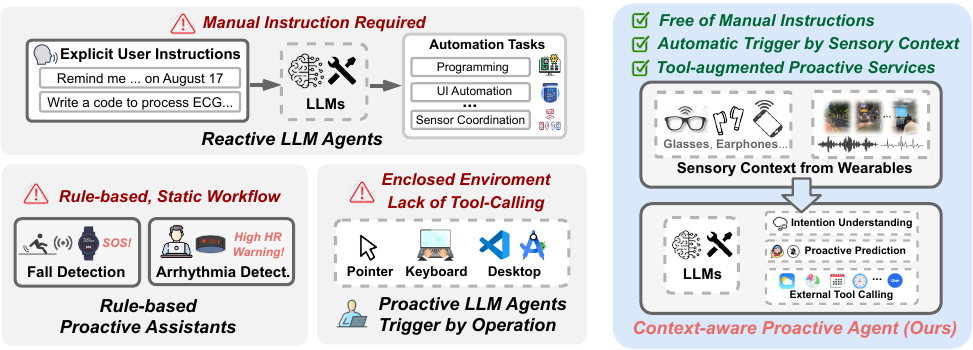}
    \caption{
Comparison with existing works. 
Reactive LLM agents require explicit user instructions to initiate tasks.
Prior proactive LLM agents focus on perceiving enclosed environments (e.g., desktop UIs) and may still require user operations (e.g., keyboard inputs) alongside direct LLM inference.
In contrast, \workname~requires no manual instructions, harnesses massive sensory contexts from the open world, and employs LLM reasoning for tool-augmented proactive services.}
    \label{fig:teaser}
\end{figure}

In this paper, we introduce \workname, the first context-aware proactive LLM agent that harnesses extensive sensory contexts for enhanced proactive services.
\workname~first employs a proactive-oriented context extraction approach to derive both sensory and persona contexts from massive sensory perceptions such as egocentric videos and audio. 
We then develop a context-aware reasoner that integrates both sensory and persona contexts for reasoning, predicts the necessity of proactive services, and calls external tools when necessary.
This reasoner is fine-tuned with reasoning traces distilled from advanced reasoning LLMs, enabling it to think before acting.
Fig.~\ref{fig:teaser_ours} shows an example scenario where the user arrives at a bus stop just after the bus has left.
\workname~can leverage this sensory context to proactively deliver useful services, such as real-time bus schedules, and determine whether alternative transportation is needed.
By harnessing sensor perceptions from hands-free, egocentric wearables (e.g., smart glasses and earphones), along with LLM reasoning, \workname~moves closer toward a more ubiquitous and proactive AI assistant.

To better examine \workname, we further introduce a new benchmark, \benchmark, for the comprehensive evaluation of context-aware proactive LLM agents.
\benchmark~contains 1,000 samples covering nine daily life scenarios, such as working and chitchat, and includes twenty external tools.
We conduct comprehensive evaluations, comparing \workname~to six baselines and testing on 13 LLMs.
Results show that \workname~achieves the state-of-the-art performance in proactive predictions and tool calling.
We summarize the contributions of this work as follows.
\begin{itemize}[leftmargin=*]
\item 

We raise the research problem of context-aware proactive agents that harness extensive sensory contexts surrounding humans to enhance the proactivity of the LLM agents and deliver tool-augmented proactive services, moving toward the vision of proactive personal assistants.

\item 
We propose \workname, the first framework for context-aware proactive LLM agents.
\workname~employs a novel context extraction method that derives sensory and persona contexts from massive sensor perceptions.
Additionally, we design a context-aware reasoner with think before action capabilities that can integrate both sensory and persona contexts for reasoning, predict the necessity of proactive services, and call external tools when necessary to assist the user.

\item 
We introduce \benchmark, the first benchmark for the comprehensive evaluation of context-aware proactive LLM agents.
Extensive evaluation on \benchmark~shows that \workname~outperforms six baselines by achieving up to 8.5\% higher accuracy for proactive predictions, 7.0\% higher F1-score for tool calling, and 6.0\% higher accuracy for tool arguments.
\end{itemize}

\section{Related Works}

\noindent\textbf{Reactive LLM-based Agents.}
Recent studies have proposed various LLM agents to perform complex tasks, such as automated web navigation~\cite{deng2023mind2web,zheng2024gpt}, software engineering \cite{yang2024swe}, personal assistant \cite{yang2025socialmind,yang2024viassist}, and household robotics \cite{ahn2022can}.
Additionally, prior research has primarily focused on enhancing the core capabilities of LLM agents, including task planning~\cite{yao2023react}, function calling~\cite{qin2023toolllm,liu2024toolace,shi2024chain}, experience reflection~\cite{zhao2024expel,shinn2023reflexion}, generalization abilities~\cite{wu2024avatar,paranjape2023art}, and multi-agent collaboration~\cite{trirat2024automl,liu2024autonomous,zhang2024chain}.
Other studies have explored the LLM agents in mobile systems, such as autonomous UI operations on smartphones~\cite{wen2024autodroid,
lee2024mobilegpt,
zhang2023appagent} and embedded programming~\cite{shen2025autoiot,englhardt2024exploring,yang2024embedgenius}.
However, although numerous frameworks and optimizations have been proposed, prior research has primarily focused on reactive LLM agents that require explicit textual instructions from users and cannot utilize the extensive contextual information from sensor perceptions on wearable devices to enable proactive assistance.

\noindent\textbf{Proactive LLM Agents.}
Proactive agents aim to autonomously initiate services based on environmental observations, without requiring explicit user instructions, evolving from early rule-based or periodic triggers~\cite{fall_detection_apple_watch} to recently proposed LLM-based approaches~\cite{zhao2025codinggenie,lu2024proactive}.
Ask-before-plan~\cite{zhang2024ask} employs re-asking strategies to proactively reduce ambiguity in a user’s instructions and enhance subsequent planning, although it still requires an initial user query.
ProAgent~\cite{zhang2024proagent} is a proactive cooperation framework among multiple robot agents, while its proactive design primarily focuses on predicting teammates’ actions in multi-agent systems rather than the user’s intention.
Recent studies, such as Proactive Agent~\cite{lu2024proactive} and CodingGenie~\cite{zhao2025codinggenie}, also propose proactive LLM agents that monitor the user interface environment on computer systems and proactively assist with tasks such as coding and writing.
However, existing work either leverages observations on computer interfaces or employs a re-asking strategy to gather more information, without utilizing the rich sensory contexts to proactively initiate services.
Moreover, prior works primarily use LLMs for direct inference rather than integrating external tools, resulting in limited proactive service functionality.

\noindent\textbf{LLM Agent Benchmark.}
A diverse and large-scale benchmark is essential for the comprehensive evaluation of LLM agents.
However, existing benchmarks primarily focus on reactive LLM agents~\cite{yang2024swe,qin2023toolllm,mialon2023gaia,deng2024mobile,liu2024toolace}, where the agent needs to take user instructions as inputs and perform task planning and tool calling.
Although a recent work~\cite{lu2024proactive} proposes ProactiveBench, it is limited to an enclosed environment, i.e., desktop UI, and does not leverage the rich contextual information from multi-modal sensors on wearable devices.
Additionally, ProactiveBench relies on direct LLM inference for responses, instead of calling diverse external tools.
Therefore, a research gap remains in developing a comprehensive benchmark for evaluating proactive LLM agents that incorporate the rich contextual information from wearable devices for proactive reasoning with tool-calling capabilities.

\section{Context-aware Proactive Agent Task}

\subsection{Task Definition}
In contrast with existing reactive LLM agents and proactive agents that rely solely on observations from desktop interfaces or direct inference, we formalize context-aware proactive LLM agents as: 
$(\mathcal{T},\mathcal{P_S},\mathcal{T_C},\mathcal{R})=\mathcal{A} (\mathcal{S}, \mathcal{P})$,
where $\mathcal{A}$ is the LLM agent, which integrates the sensory perceptions $\mathcal{S}$ and persona context $\mathcal{P}$ as input.
Here $\mathcal{S}$ contains sensor perceptions from multi-modal wearables such as smart glasses and earphones, including egocentric video $\mathcal{S_V}$, audio $\mathcal{S_A}$, and smartphone notification $\mathcal{N}$.
We denote the sensory context $\mathcal{C}$ as the implicit cues within the raw sensory perceptions $\mathcal{S}$ that help determine the need for proactive services. 
We also formalize that the agent should consider user personas $\mathcal{P}$ for proactive reasoning, including a person's identity, preferences, and historical behaviors.
Using these contexts, the agent generates $(\mathcal{T},\mathcal{P_S},\mathcal{T_C},\mathcal{R})$, where $\mathcal{T}$ denotes the explicit thought traces.
$\mathcal{P_S}$ denotes the proactive score, which triggers proactive services when $\mathcal{P_S} \geq \theta$.
Here, $\theta$ denotes the threshold for initiating proactive services and is a user-adjustable parameter reflecting the user’s sensitivity to such services.
$\mathcal{T_C}$ is the planned tool chains that LLM agents should call in sequence, where $\mathcal{T_C} = {(t_i, a_i)}_{i=1}^{N},t_i\in \mathbf{T}$, with $t_i$ as each tool to be called and $a_i$ as the corresponding arguments.
$\mathbf{T}$ is the tool set that the agent can use.
$\mathcal{R}$ is the agent's final response, summarizing the sensory context, persona context, reasoning traces, and tool results.
Note that proactive assistance is only initiated when $\mathcal{P_S} \geq \theta$, otherwise the agent does not disturb the user.

\subsection{Task Construction}
\label{sec:dataset_construction}
Recognizing the shortcomings of existing LLM agent benchmarks, we present \benchmark, the first benchmark designed to evaluate context-aware proactive LLM agents.

\noindent\textbf{Design Choices.}
Our dataset includes the following key features: 1) 
\textit{Sensory Context.}
Our dataset contains sensory context obtained from wearables (e.g., smart glasses and earphones), which capture shared perceptions of the user \textit{ubiquitously}.
This hands-free captured sensory context is more suitable for proactive agents as it can reduce the user's physical and cognitive workload, aligning with the mission of proactive agents.
2). \textit{Persona Context}. 
We incorporate diverse personas to support more comprehensive and personalized scenarios for proactive services.
3) \textit{Proactive Assistance with Tool Calling.}
The dataset targets tool-using LLM agents that map the contexts to proactive assistance by utilizing multiple external tools to generate more informative responses, rather than direct inference.

\noindent\textbf{Formulation and Exemplar Design.}
Each sample in our dataset contains seven parts: $(\mathcal{S},\mathcal{C}, \mathcal{P}, \mathcal{T}, \mathcal{P_S}, \mathcal{T_C}, \mathcal{R})$.
Next, we introduce the design of initial exemplars.

\textit{Multi-dimensional Context Information.}
Annotators first write textual descriptions of their egocentric perceptions, including what they see, hear, and any mobile device notifications, for both proactive and non-proactive scenarios that they encounter in daily life.
This sensory perception can be captured from an egocentric perspective using various wearable devices.
The context information contains the visual context $\mathcal{C_V}$, the acoustic context $\mathcal{C_A}$, and the notifications on the smartphone $\mathcal{N}$.
Annotators also summarize them into contextual information $\mathcal{C}$, providing a comprehensive description of the user's current conditions.
Annotators write the user personas $\mathcal{P}$ for the sample if necessary.
The persona can include any information about a person's preferences or identity.

\textit{Proactive Score with Planned Tool Chains.}
Next, annotators are instructed to analyze the current context and assign a proactive score $\mathcal{P_S}$.
We define $\mathcal{P_S}$ on a scale from 1 to 5, where 1 means that no proactivity is required and 5 means a high level of proactivity.
Annotators also receive a tool set $T$ that includes the usable tools, tool names, tool descriptions, arguments, and formats predefined by the developers.
Details are in the Appendix~\ref{sec:tool_def}.
For samples identified as requiring proactivity, we request annotators to further label the planned tool chains $\mathcal{T_C}$, specifying the external tools that agents should use. 
If $\mathcal{P_S}=1$ or $2$, both $\mathcal{T_C}$ and $\mathcal{R}$ are \texttt{None}, as there is no need for proactivity.

We instruct the annotators to create samples spanning nine everyday scenarios, ranging from work to chitchat.
We ask annotators to document their thought processes, including their analysis of the current context, their rationale for assigning the proactive score, and the planned tool chains.
Each annotator also cross-reviewed the samples produced by others, evaluating both the format and plausibility to avoid overproactivity and ensure the correctness of annotations. 
Through this process, we acquire 200 human-created exemplars to serve as the seed dataset.

\noindent\textbf{Automated Diversification Pipeline.}
Relying solely on manual efforts to scale the dataset presents challenges, as scenarios and contextual information created by humans may lack diversity and generalizability. 
Moreover, human fatigue during annotation can introduce bias, potentially compromising the dataset's quality. 
Therefore, we develop an automated diversification pipeline to use LLMs for data generation, producing a large-scale dataset with diverse samples.

\textit{Information Source.}
We first prepare several resources to help LLMs generate synthetic data, including the tool set (Appendix~\ref{sec:tool_def}), an extensive persona pool, and the initial exemplars. The personas in our pool are sourced from \cite{ge2024scaling}, which includes one billion individual identities and preferences.

\textit{Generation with Verification.}
Next, we prompt LLMs to generate diverse samples by utilizing the initial exemplars, tool set, and persona pool for reference. 
We employ two strategies during generation: scenario-aware and proactive score-aware. 
In the first strategy, we group the seed dataset by scenarios and instruct LLMs to generate samples based on specific scenarios within the nine categories.
In the second strategy, LLMs are prompted to generate samples based on a specific proactive score.
Details are in the Appendix~\ref{sec_prompts}.
After generation, annotators first evaluate the context and annotations for rationality.
Next, we execute a script to verify the correctness of the data format and tool arguments.
We perform several iterations of the above process to obtain \benchmark.

\begin{figure}[htb]
\vspace{-5pt}
    \centering
    \begin{subfigure}{0.19\columnwidth}
        \centering
\includegraphics[width=1\textwidth]{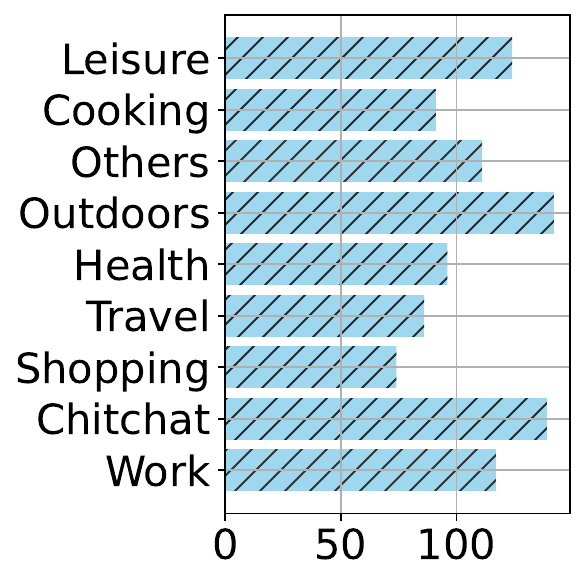}
        \vspace{-15pt}
        \caption{Scenarios.}
\label{fig:dataset_scenario_category}
    \end{subfigure}
    \begin{subfigure}{0.19\columnwidth}  
        \centering 
\includegraphics[width=1\textwidth]{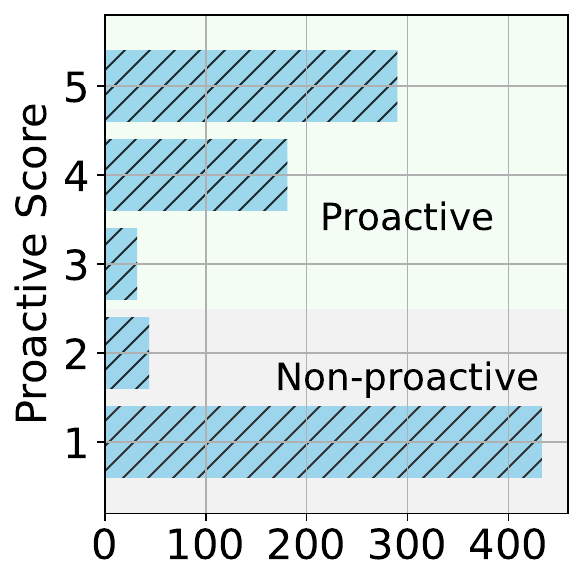}
        \vspace{-15pt}
        \caption{Proactive score.}   \label{fig:dataset_proactive_score}
    \end{subfigure}
    \begin{subfigure}{0.19\columnwidth}  
        \centering 
\includegraphics[width=1\textwidth]{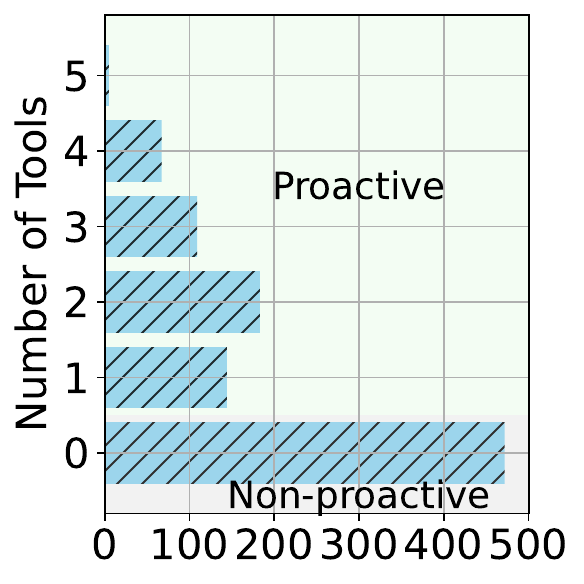}
        \vspace{-15pt}
        \caption{Tool number.}   \label{fig:dataset_tool_number}
    \end{subfigure}
    \begin{subfigure}{0.4\columnwidth}
        \centering
\includegraphics[width=0.98\textwidth]{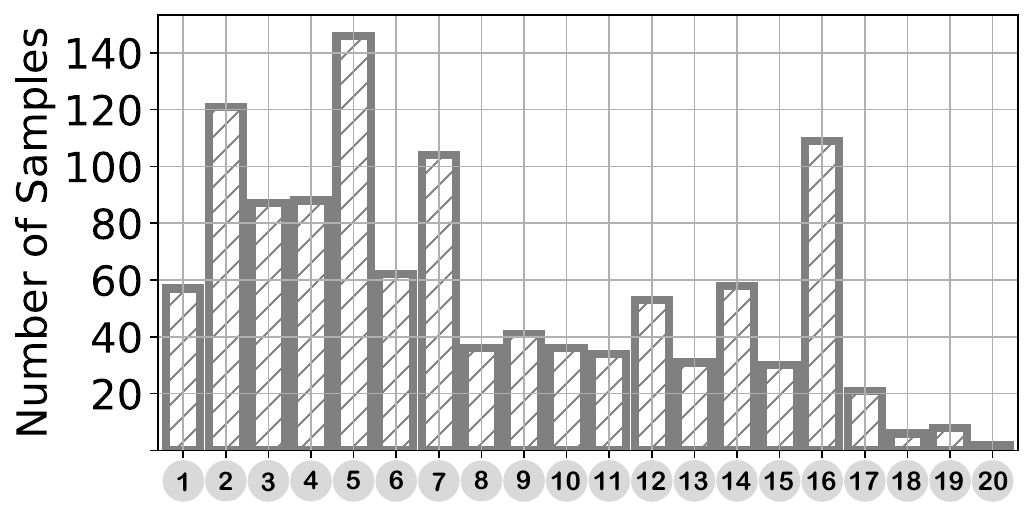}
        \vspace{-5pt}
        \caption{Tool distribution.}
    \end{subfigure}
    \caption{
Statistics of \benchmark, including the sample distribution across different scenarios, proactive scores, and the number and types of tools. In subfigures (a)–(c), the x-axis shows the number of samples, whereas in (d) it denotes the tool index.}
\label{fig:dataset_statistics}
\end{figure}

\noindent\textbf{Consideration of Sensor Data.}
We also collect raw sensor data from copyright-free internet platforms~\cite{Pexels} to pair with the textual contextual information in \benchmark.
Specifically, we first randomly select samples from \benchmark, and scrape the videos from Pexels~\cite{Pexels} based on the textual descriptions of the visual context information. 
Note that we exclusively collect videos captured from an egocentric perspective.
Additionally, for samples with audio conversations, we self-collect both video and audio to align with the textual context information.
Finally, we obtain the \benchmarklite, consisting of 300 human-verified samples with raw sensor data.

\noindent\textbf{Dataset Statistics.}
Fig.~\ref{fig:dataset_statistics} shows the statistics of our dataset.
We collected 1,000 samples for \benchmark~and 300 samples for \benchmarklite.
Our dataset covers 9 daily life scenarios and includes 20 tool types, with each sample potentially involving the use of up to five tools.
We provide more details on the dataset and tool definitions in the Appendix~\ref{dataset_details}.
\section{\workname~Framework}
\label{sec:framework}

\begin{figure}[htb]
    \centering
    \vspace{-10pt}
    \includegraphics[width=\linewidth]{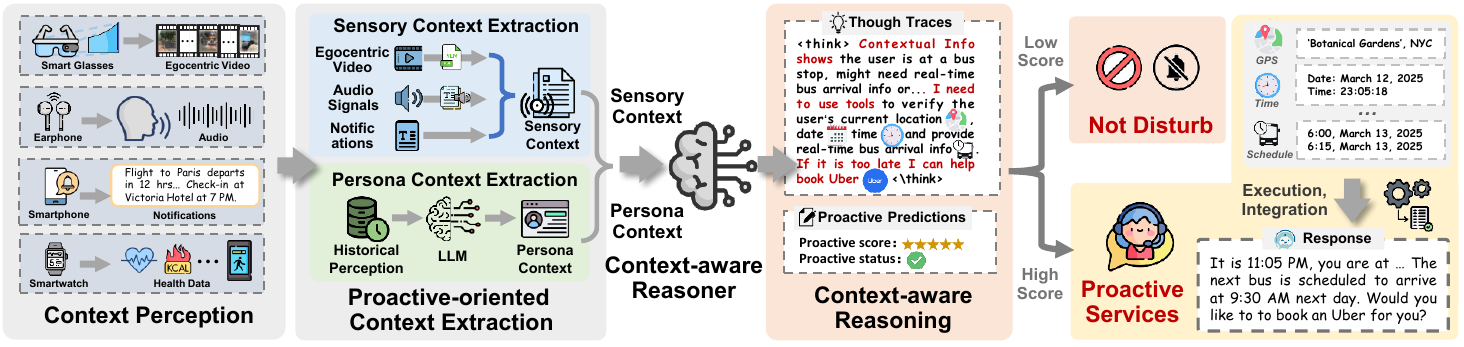}
    \vspace{-15pt}
    \caption{
    Overview of \workname. \workname~extracts sensory context from massive sensor perceptions.    
    Then it integrates both sensory and persona contexts into LLM reasoning, generating thought traces, proactive predictions, and calling external tools for proactive services when necessary.}
    \label{fig:system}
    \vspace{-10pt}
\end{figure}

This section presents the framework of \workname, introducing how it utilizes the massive sensory contexts for tool-augmented proactive LLM agent services.
Fig.~\ref{fig:system} shows the overview of \workname. First, \workname~extracts proactive-oriented contexts from multi-modal sensory perceptions. 
Next, \workname~integrates these contexts for tool-augmented proactive services.

\subsection{Proactive-oriented Context Extraction}
Previous studies focus on extracting sensory contexts and use LLMs to summarize insights~\cite{post2025contextllm}. However, relying solely on these sensory contexts can lead to inferior proactive predictions.
Therefore, \workname~employs a proactive-oriented context extraction method.
In \workname, contexts comprise two types: \textit{sensory context} and \textit{persona context}.
The sensory context includes insights for the user's surroundings and actions, which are crucial for inferring user intent.
The persona context encompasses user personal information, including past behaviors, preferences, and identity, which helps LLM agents to determine the need and urgency for proactive assistance.
This subsection provides details on how \workname~extracts these contexts from extensive sensory perceptions.

\noindent\textbf{Sensory Context.}
\workname~first employs Vision Language Models (VLMs)~\cite{liu2023visual} to transform raw egocentric videos into visual contexts $\mathcal{C_V}$.
Although existing VLMs can generate video descriptions, they often produce overly simplistic descriptions that overlook crucial cues for understanding user intent or overly detailed, redundant insights, both of which can hinder the LLM agent’s proactive predictions.
Thus, instead of using zero-shot VLMs, \workname~employs in-context learning~(ICL) to generate proactive-oriented visual contexts $\mathcal{C_V}$.
\workname~also employs the speech recognition model to generate audio contexts $\mathcal{C_A}$.
Details and prompts for sensory context extraction are in the Appendix~\ref{sec_prompts}.
Finally, \workname~integrate these contexts into the final context information $\mathcal{C}=[\mathcal{C_V},\mathcal{C_A},\mathcal{N}]$, which includes visual contexts $\mathcal{C_V}$, audio contexts $\mathcal{C_A}$, and textual information from smartphone notifications $\mathcal{N}$, such as calendar events and hotel reservations.

\noindent\textbf{Persona Context.}
Since the need for assistance highly depends on the user's personal preferences, \workname~also integrates persona context into its reasoning. 
In this work, we use persona contexts within \benchmark~for experiments. 
In practice, these contexts can be continuously updated by utilizing LLMs to extract insights from historical sensory data like daily conversations~\cite{yang2025socialmind}.

\subsection{Context-aware Proactive Reasoning}
While existing LLM agents can handle complex tasks based on explicit user instructions~\cite{liu2025tasksense,zheng2024gpt,cheng2024seeclick}, they face challenges when processing sensory contexts and correctly mapping them to the appropriate tools for proactive services.
Next, we will introduce the context-aware reasoner in \workname.

\noindent\textbf{Context-aware Reasoner.}
\workname~employs a context reasoner $\mathcal{A_S}$ to reason over the generated contexts and provide proactive services: $(\mathcal{T},\mathcal{P_S},\mathcal{T_C})=\mathcal{A_S}(\mathcal{C},\mathcal{P})$.
The context reasoner is an LLM that integrates both sensory context $\mathcal{C}$ and personas $\mathcal{P}$ as input to generate thought traces $\mathcal{T}$, proactive scores $\mathcal{P_S}$, and tool chains $\mathcal{T_C}$.
We enable \workname~to perform think-before-act reasoning by distilling traces from advanced LLMs (e.g., Claude-3.7-Sonnet~\cite{Claude-3.7-Sonnet}) and constructing a CoT-based~\cite{wei2022chain} fine-tuning dataset.
During inference, once $\mathcal{P_S}\geq \theta$, \workname~will initiate the proactive services.
Additionally, \workname~generates tool chains $\mathcal{T_C}$ for enhanced proactive services.
\workname~will execute the planned tools sequentially and integrate their results with the sensory context, persona context, and thought traces into the LLM to generate final responses.

\noindent\textbf{Training Scheme.}
We use supervised fine-tuning (SFT) with CoT to train the context reasoner in \workname.
Specifically, we construct the SFT dataset $\mathcal{D}_{SFT}=\{(\mathcal{X},\mathcal{T},\mathcal{Y})\}$.
Here, $\mathcal{X}$ contains the sensory context $\mathcal{C}$ and persona context $\mathcal{P}$.
The thought traces $\mathcal{T}$ divided by $\texttt{<think>}$ and $\texttt{</think>}$, are distilled from advanced LLMs~\cite{Claude-3.7-Sonnet}, enabling \workname~to ``think before acting'', generating explicit thought traces before proactive predictions and tool calls.
The output $\mathcal{Y}$ contains proactive scores $\mathcal{P_S}$ and planned tool chains $\mathcal{T_C}$.
\section{Experiments}
\subsection{Experimental Setup}
\label{sec:implement_details}

\noindent\textbf{Implementation Details.}
Our experiments are conducted using 8 A6000 GPUs. For SFT, we use the AdamW optimizer with a learning rate of 0.0001 and apply LoRA techniques during model training.  
We set the LoRA rank to 8 and use a cosine scheduler with a 10\% warmup ratio, training for 5 epochs.
For ICL-based baselines, we randomly select 10 samples from the dataset as demonstrations included in the prompt.
We randomly split the dataset into 60\% training and 40\% testing in our experiments.

\noindent\textbf{Metrics.}
We employ two categories of metrics to evaluate the performance of context-aware proactive LLM agents, including proactive prediction and tool calling. Details of each metric are as follows.

\begin{itemize}[leftmargin=*]
\item
\textbf{Proactive Prediction.}
We first evaluate the agent's ability to accurately determine the need for initiating proactive services.
Specifically, we use four metrics to assess proactive prediction performance, including the accuracy of proactive predictions (\textbf{Acc-P}), missed detections (\textbf{MD}), false detections (\textbf{FD}), and the root mean square error (\textbf{RMSE}) between predicted proactive scores and ground-truth. 
Acc-P, MD, and FD are commonly used in the existing work~\cite{lu2024proactive}, while RMSE provides a finer-grained evaluation of the performance of predicted proactive scores.
 
\item
\textbf{Tool Calling.}
To evaluate the agent's tool calling performance, we follow existing works~\cite{abdelaziz2024granite,basu2024api} and use standard metrics such as \textbf{Precision}, \textbf{Recall}, and \textbf{F1-score} to compare the tool names in the predicted tool set with those in the ground-truth tool set.
We also use \textbf{Acc-Args} to evaluate whether the proactive agent can correctly generate the structured data for tool calls, including the tool names and arguments. If an argument of any tool is incorrect, the entire sample is considered incorrect.
For Acc-Args, we calculate the accuracy only for the correctly predicted tools to ensure a fair comparison of different approaches.
\end{itemize}

\noindent\textbf{Baselines.}
We compare \workname~with several baselines, including Proactive Agent~\cite{lu2024proactive}, vanilla ICL, CoT, ICL-P, ICL-All, vanilla SFT, and SFT-P.
For the Proactive Agent, we follow~\cite{lu2024proactive} and modify the task instructions in the system prompt to adapt to the proactive agent task.
For the vanilla ICL, we use few-shot demonstrations with only sensory contexts. 
For CoT, we include both sensory contexts and thought traces. 
For ICL-P, we include sensory contexts and personas, and for ICL-All, we incorporate sensory contexts, thought traces, and personas into the prompt.
Vanilla SFT uses sensory contexts for fine-tuning. SFT-P uses both sensory and persona contexts.
We conduct experiments on 13 LLMs, comprising \textbf{(1) proprietary LLMs} including GPT-4o~\cite{hurst2024gpt} GPT-3.5~\cite{ye2023comprehensive}, GPT-o3, GPT-o4-mini, and Claude Sonnet 4, \textbf{(2) open-source LLMs} including Llama-3.1-70B-Instruct~\cite{grattafiori2024llama} and Qwen2.5-72B-Instruct~\cite{yang2024qwen2}, and \textbf{(3) small LLMs} including Llama-3.1-8B-Instruct, Qwen2.5-7B-Instruct, Qwen2.5-3B-Instruct, Qwen2.5-1.5B-Instruct, DeepSeek-R1-Distill-Qwen-7B~\cite{deepseekai2025deepseekr1incentivizingreasoningcapability}, and DeepSeek-R1-Distill-Qwen-1.5B.
Details of the baselines and implementation are in the Appendix~\ref{sec_prompts}.

\subsection{Results on Benchmarks}
\begin{figure}[h]
    \centering
\includegraphics[width=\textwidth]{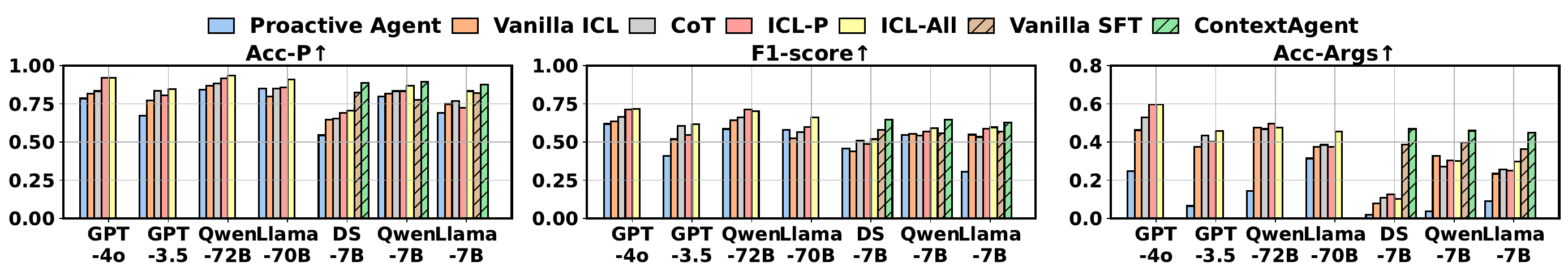}
    \caption{Main results on ContextAgentBench.
    `DS' refers to `DeepSeek'.
    }
    \label{fig:main_results_benchmark}
\end{figure}

\noindent\textbf{Quantitative Results on \benchmark.}
Tab.~\ref{tab:main_results_benchmark} shows the overall performance of \workname~on \benchmark.
Results show that when using Llama3.1-8B-Instruct as the base LLM, \workname~consistently achieves the highest performance across all metrics, with increases of 8.5\% in Acc-P, 7.0\% in F1-score, and 6.0\% in Acc-Args.
Fig.~\ref{fig:main_results_benchmark} shows that \workname~can achieve performance comparable to or even exceeding baselines that employ 70B-scale LLMs and proprietary LLMs, with metrics such as Acc-P (-1.5\%), F1-score (-3.0\%), and Acc-Args (+6.6\%).
Due to space constraints, Fig.~\ref{fig:main_results_benchmark} shows only three key metrics. 
See Tab.~\ref{tab:main_results_benchmark_all} in Appendix~\ref{full_results_on_benchmark} for full comparison.

\begin{table}[t]
\setlength\tabcolsep{5pt}
\centering
\scriptsize
\caption{
Main results on \benchmark.
}
\label{tab:main_results_benchmark}
\begin{threeparttable}
\begin{tabular}{c|c|c|c|c|c|c|c|c|c}
\toprule
\multirow{2}{*}{\textbf{Model}}     & \multirow{2}{*}{\textbf{Method}} &\multicolumn{4}{c|}{\textbf{Proactive Predictions}}  & \multicolumn{4}{c}{\textbf{Tool Calling}}  \\ \cmidrule(l){3-10} 
&  & \textbf{Acc-P$^{\uparrow}$} &\textbf{MD.$^{\downarrow}$}
&\textbf{FD.$^{\downarrow}$}  & \textbf{RMSE$^{\downarrow}$}   & \textbf{Precision$^{\uparrow}$}  & \textbf{Recall$^{\uparrow}$}   & \textbf{F1-score$^{\uparrow}$}  & \textbf{Acc-Args$^{\uparrow}$}  \\ 

\midrule
\multirow{7}{*}{Llama-3.1-8B-Ins}     & Proactive Agent& 0.676& 0.017& 0.306 & 1.915 & 0.397 &0.290 &0.318 & 0.081 \\
& Vanilla ICL   & 0.742& 0.224& 0.033 & 1.853& 0.608 & 0.533 & 0.552& 0.269\\
& CoT  & 0.699& 0.278& 0.023 & 1.960& 0.590 &0.539  & 0.551& 0.209\\
& ICL-P  & 0.742& 0.242& 0.015 &1.922 &0.608  & 0.553 & 0.567& 0.262\\
& ICL-All  & 0.757& 0.229& 0.012 &1.872 & 0.631 & 0.565 & 0.582& 0.270\\
& Vanilla SFT  & 0.813& 0.068& 0.117 & 1.572& 0.609 & 0.581 & 0.580&0.405\\

&\cellcolor{mygray}\textit{\workname}  & 
\cellcolor{mygray}0.874& \cellcolor{mygray}0.030&  \cellcolor{mygray}0.095& \cellcolor{mygray}1.408& \cellcolor{mygray}0.660& \cellcolor{mygray}0.627& \cellcolor{mygray}0.626& \cellcolor{mygray}0.448\\

\midrule
\multirow{7}{*}{DeepSeek-R1-7B}     & Proactive Agent& 0.544& 0.411& 0.044 & 3.093& 0.467 & 0.454 & 0.457&0.019 \\
& Vanilla ICL   & 0.646& 0.248& 0.105 & 2.568& 0.457 & 0.433 & 0.437& 0.078\\
& CoT  & 0.653& 0.319& 0.027 & 2.760& 0.528 & 0.501 & 0.507& 0.109\\
& ICL-P  & 0.690&  0.227 & 0.081& 2.466 & 0.518& 0.479& 0.486& 0.126\\
& ICL-All  & 0.704& 0.268& 0.0272 & 2.540& 0.545 & 0.510 & 0.518& 0.103\\
& Vanilla SFT  & 0.823& 0.068&  0.108& 1.630& 0.621&  0.570& 0.579& 0.386\\
&\cellcolor{mygray}\textit{\workname}  & \cellcolor{mygray}0.888& \cellcolor{mygray}0.027&  \cellcolor{mygray}0.085& \cellcolor{mygray}1.319& \cellcolor{mygray}0.676& \cellcolor{mygray}0.648& \cellcolor{mygray}0.647& \cellcolor{mygray}0.468\\

\midrule
\multirow{7}{*}{Qwen2.5-7B-Ins}     & Proactive Agent& 0.799& 0.136& 0.064 & 2.038& 0.578 & 0.536 & 0.546& 0.038\\
& Vanilla ICL   & 0.816& 0.088& 0.095 & 1.752& 0.590 & 0.545 & 0.553& 0.326\\
& CoT  & 0.833& 0.085& 0.081 & 1.790& 0.585 & 0.527 & 0.541& 0.272\\
& ICL-P  & 0.833& 0.091& 0.074 & 1.819& 0.610 & 0.556 & 0.568& 0.303\\
& ICL-All  &0.867 & 0.088& 0.044 & 1.721& 0.635 & 0.577 & 0.591& 0.301\\
& Vanilla SFT  & 0.775& 0.088& 0.136 & 1.774& 0.589 & 0.551 & 0.558&0.398\\
& \cellcolor{mygray}\textit{\workname}  &\cellcolor{mygray}0.894 &\cellcolor{mygray}0.013 & \cellcolor{mygray}0.091 & \cellcolor{mygray}1.264& \cellcolor{mygray}0.672 & \cellcolor{mygray}0.644 & \cellcolor{mygray}0.645& \cellcolor{mygray}0.459\\

\bottomrule
\end{tabular}
\end{threeparttable}
\vspace{-1.5em}
\end{table}

\begin{figure}[!h]
    \centering
\includegraphics[width=\textwidth]{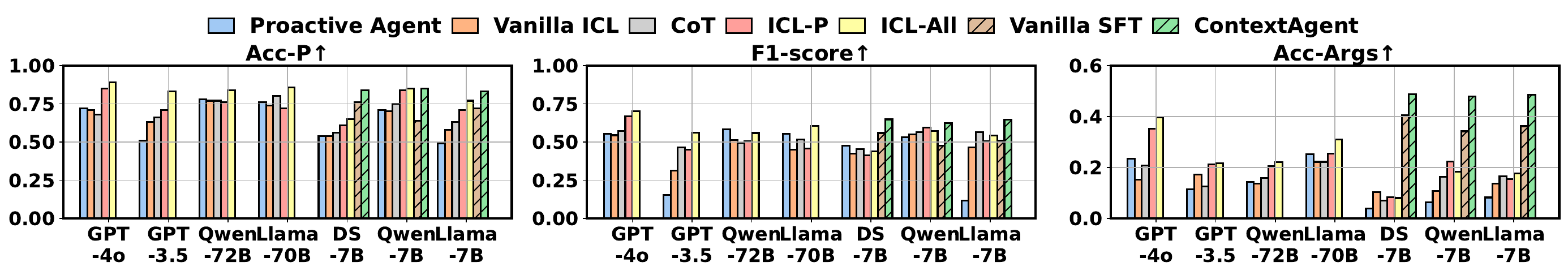}
\vspace{-15pt}
    \caption{Main results on \benchmarklite. 
    }    \label{fig:main_results_benchmark_lite}
\vspace{-10pt}
\end{figure}

\noindent\textbf{Quantitative Results on \benchmarklite.}
Fig.~\ref{fig:main_results_benchmark_lite} and Tab.~\ref{tab:main_benchmark_lite} show the performance of \workname~on \benchmarklite.
The results indicate that both \workname~and the baselines exhibit slight performance degradation. However, \workname~still achieves the highest performance across all metrics compared to the baselines. When using Qwen2.5-7B-Instruct as the base LLM, \workname~achieves improvements of 6.2\% Acc-P, 3.0\% F1-score, and 7.6\% Acc-Args, over the best baseline. 
It can even achieve comparable and even higher performance than baselines using 70B-scale and proprietary LLMs.
Complete results are provided in Tab.~\ref{tab:main_results_benchmark_lite} within Appendix~\ref{full_results_on_benchmark}.

\begin{figure}[t]
  \centering
  \vspace{-5pt}
\includegraphics[width=1\linewidth]{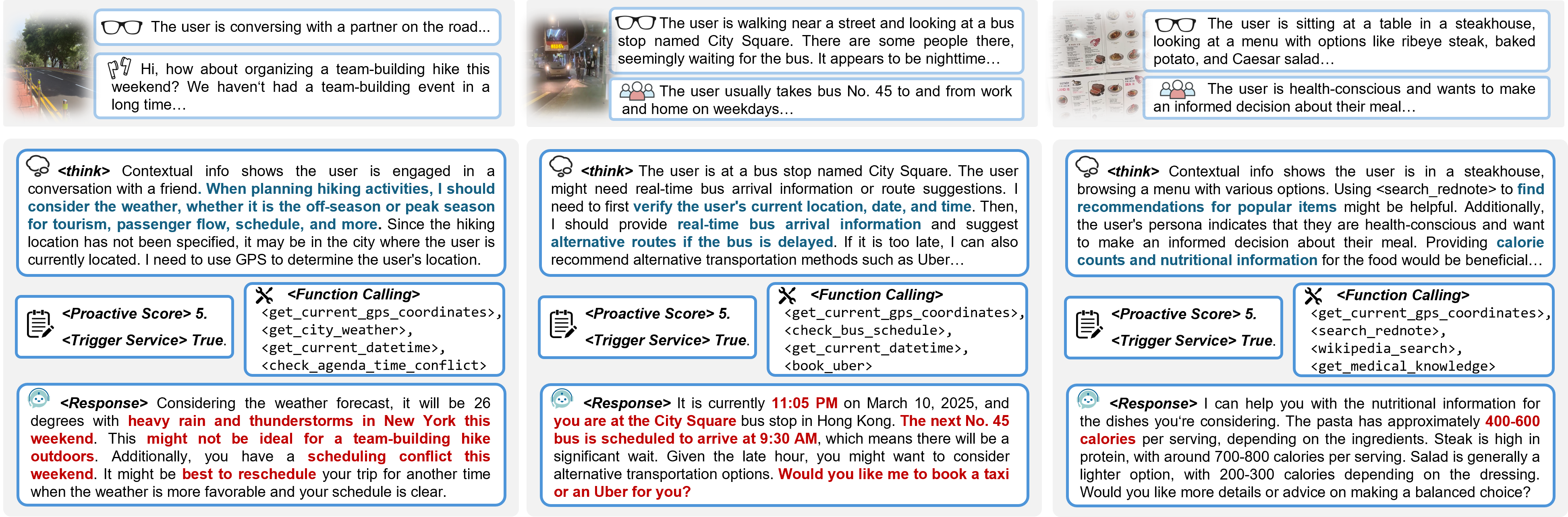}
\vspace{-15pt}
\caption{Qualitative results of \workname~in proactive cases, showing sensory and persona contexts, and \workname's thought traces, proactive predictions, tool calls, and final responses.}
  \label{fig:qualitative_results_overall_proactive}
  \vspace{-1em}
\end{figure}

\begin{figure}
  \centering
  \vspace{-10pt}
\includegraphics[width=1\linewidth]{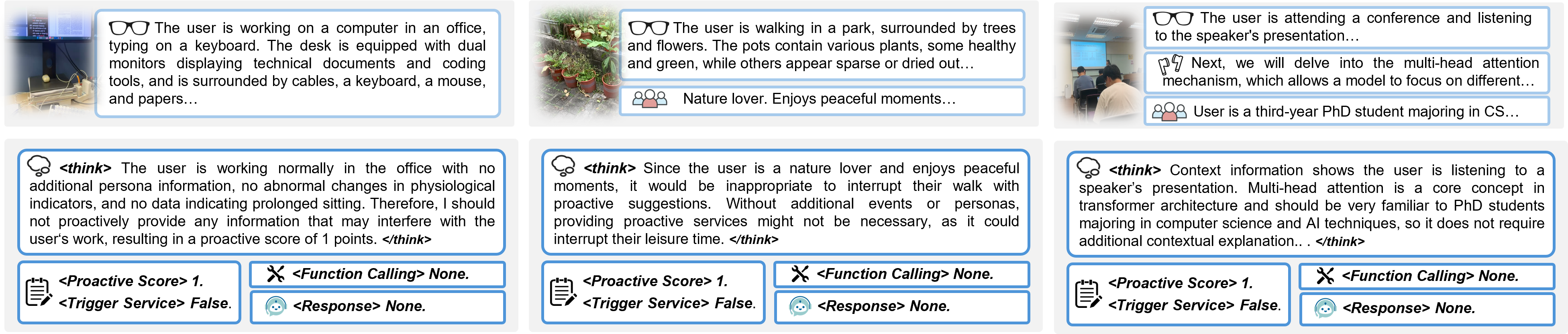}
\caption{Qualitative results of \workname~in non-proactive cases.}
  \label{fig:qualitative_results_overall_noproactive}
\end{figure}

\noindent\textbf{Qualitative Results.}
Fig.~\ref{fig:qualitative_results_overall_proactive} and Fig.~\ref{fig:qualitative_results_overall_noproactive} show the qualitative results of \workname, including both proactive and non-proactive cases.
\textbf{First}, \workname~integrates both sensory and persona contexts for reasoning and offers appropriate proactive services.
\textbf{Second}, it can further map these contexts to planned tool chains such as the weather and agenda checker, and integrate external knowledge for enhanced proactive service.
Fig.~\ref{fig:qualitative_results_overall_proactive} shows that \workname~can employ GPS, datetime tools, bus schedule checkers, and ride-booking apps like Uber when the user is approaching a bus station.
Furthermore, during casual conversations involving proposed outdoor activities, \workname~uses tools such as a weather checker, a datetime tool, and an agenda checker to proactively assist the user in evaluating feasibility and making informed decisions.
\textbf{Third}, \workname~can leverage persona context to generate more personalized proactive predictions.
For instance, Fig.~\ref{fig:qualitative_results_overall_proactive} shows that for health-conscious individuals deciding what to order at a restaurant, \workname~proactively offers food-related health information and suggestions.
Fig.~\ref{fig:qualitative_results_overall_noproactive} shows that the persona contexts also help \workname~to determine not to disrupt the users.
See Appendix~\ref{more_results} for more qualitative results.

\begin{table}[!t]
\setlength\tabcolsep{5pt}
\centering
\scriptsize
\caption{
Main results on \benchmarklite.
}
\label{tab:main_benchmark_lite}
\begin{threeparttable}
\begin{tabular}{c|c|c|c|c|c|c|c|c|c}
\toprule
\multirow{2}{*}{\textbf{Model}}     & \multirow{2}{*}{\textbf{Method}} &\multicolumn{4}{c|}{\textbf{Proactive Predictions}}  & \multicolumn{4}{c}{\textbf{Tool Calling}}  \\ \cmidrule(l){3-10} 
&  & \textbf{Acc-P$^{\uparrow}$} &\textbf{MD.$^{\downarrow}$}
&\textbf{FD.$^{\downarrow}$}  & \textbf{RMSE$^{\downarrow}$}   & \textbf{Precision$^{\uparrow}$}  & \textbf{Recall$^{\uparrow}$}   & \textbf{F1-score$^{\uparrow}$}  & \textbf{Acc-Args$^{\uparrow}$}  \\ 

\midrule
\multirow{7}{*}{Llama3.1-8B-Ins}     & Proactive Agent& 0.490& 0.000& 0.510 & 2.469& 0.200 & 0.088 & 0.117&0.082 \\
& Vanilla ICL   & 0.580& 0.320& 0.100 & 2.623& 0.500 & 0.450 & 0.466& 0.136\\
& CoT  & 0.630& 0.360& 0.010 & 2.306& 0.595 & 0.553 & 0.564& 0.166\\
& ICL-P  & 0.710&  0.210 & 0.080& 2.315 & 0.535& 0.495& 0.506& 0.155\\
& ICL-All  & 0.770& 0.170& 0.060 & 1.757& 0.598 & 0.526 & 0.543& 0.177\\
& Vanilla SFT  & 0.720& 0.120& 0.160 & 1.959& 0.536 & 0.497 & 0.508&0.362\\
& SFT-P  & 0.734& 0.115&  0.151& 1.980& 0.555&  0.582& 0.552& 0.353\\
&\cellcolor{mygray}\textit{\workname}  & \cellcolor{mygray}0.830& \cellcolor{mygray}0.070&  \cellcolor{mygray}0.100& \cellcolor{mygray}1.510& \cellcolor{mygray}0.687& \cellcolor{mygray}0.637& \cellcolor{mygray}0.645& \cellcolor{mygray}0.486\\

\midrule
\multirow{7}{*}{Qwen2.5-7B-Ins}     & Proactive Agent& 0.710& 0.210& 0.080 &2.328 & 0.575 & 0.515 & 0.532& 0.064\\
& Vanilla ICL   & 0.700& 0.280& 0.020 & 2.596& 0.595 & 0.533 & 0.550& 0.107\\
& CoT  & 0.750& 0.230& 0.020 &2.306 & 0.630 &0.541  &0.564 & 0.163\\
& ICL-P  & 0.840& 0.080& 0.080 & 1.783& 0.656 & 0.570 & 0.595& 0.224\\
& ICL-All  & 0.850& 0.100& 0.050 & 1.780& 0.615 & 0.565 & 0.573& 0.183\\
& Vanilla SFT & 0.640& 0.190& 0.170 & 2.206& 0.520 & 0.457 & 0.476& 
0.342 \\
& SFT-P  & 0.774& 0.083&  0.143& 1.790& 0.481&  0.495& 0.473& 0.374\\
& \cellcolor{mygray}\textit{\workname}  &\cellcolor{mygray}0.850 &\cellcolor{mygray}0.050& \cellcolor{mygray}0.100& \cellcolor{mygray}1.403& \cellcolor{mygray}0.667& \cellcolor{mygray}0.615& \cellcolor{mygray}0.624& \cellcolor{mygray}0.479\\

\bottomrule
\end{tabular}
\end{threeparttable}
\end{table}

\subsection{Ablation Study and Discussion}
\label{sec:ablation_study}

\textbf{Impact of Modalities.}
We evaluate \workname~on \benchmarklite~to assess sensitivity to missing modalities.
Tab.~\ref{tab:sensor_miss} shows that when vision or audio is missing, Acc-P decreases by up to 17.9\% and F1-score decreases by up to 23.3\%.
The results show that both modalities are critical for the context-aware proactive agent, with missing vision having a larger impact than missing audio.

\begin{table}[h]
\centering
\scriptsize
\caption{
Performance with missing sensors. 
``w/o vision'' and ``w/o audio'' denote inference without visual or audio contexts in \workname. The base model is DeepSeek-R1-7B.}
\label{tab:sensor_miss}
\begin{threeparttable}
\begin{tabular}{@{}c|c|c|c|c|c|c|c|c@{}}
\toprule
\multirow{2}{*}{\textbf{Settings}} &\multicolumn{4}{c|}{\textbf{Proactive Predictions}}  & \multicolumn{4}{c}{\textbf{Tool Calling}} \\ \cmidrule(l){2-9} 
 &  \textbf{Acc-P$^{\uparrow}$} &\textbf{MD.$^{\downarrow}$}
&\textbf{FD.$^{\downarrow}$}  & \textbf{RMSE$^{\downarrow}$}   & \textbf{Precision$^{\uparrow}$}  & \textbf{Recall$^{\uparrow}$}   & \textbf{F1-score$^{\uparrow}$}  & \textbf{Acc-Args$^{\uparrow}$}\\ 

\midrule
w/o vision   & 0.709 & 0.129& 0.163 & 2.070 & 0.424& 0.422 &0.414 &0.163\\
w/o audio  & 0.720 & 0.173& 0.106 & 2.020 & 0.501& 0.501 & 0.493& 0.212\\
Full  & 0.888& 0.027& 0.085& 1.310& 0.676& 0.648& 0.647& 0.468\\

\bottomrule
\end{tabular}
\end{threeparttable}
\end{table}
\vspace{-5pt}

\textbf{Sensory Context Perception.}
We conduct experiments using the zero-shot Qwen-2.5-VL as the VLM for sensory context extraction in \workname.
Tab.~\ref{tab:ablation_study_benchmark} shows that this causes \workname~to decrease in Acc-P, F1-score, and Acc-Args by 3.0\%, 3.3\%, and 1.9\%, respectively. 
We observe that the context generated by zero-shot VLM lacks key proactive-oriented cues, such as simply describing the user tying their shoe while sitting on the floor. In contrast, our sensory context extraction module captures detailed scenario information about the gym and specific fitness equipment, providing deeper insight into the user's conditions and intents and resulting in higher performance.

\vspace{-5pt}
\textbf{Persona Context.}
We also conduct experiments to study the impact of user personas by removing them during both the training and testing phases.
Tab.~\ref{tab:ablation_study_benchmark} shows that removing personas consistently leads to significant performance drops, with Llama-3.1-8B-Ins experiencing decreases of up to 9.0\% in Acc-P, 12.3\% in F1-score, and 12.6\% in Acc-Args. 
Results show that personas are crucial for the proactive agent task, impacting both proactive predictions and tool-calling capabilities.

\textbf{Thought Traces.}
We also investigate the impact of thought traces on the context-aware proactive agent.
We observe that integrating those thought traces can significantly improve ICL performance.
Tab.~\ref{tab:main_results_benchmark} shows that ICL-All achieves up to 20.1\% improvement in Acc-P compared to ICL-P, which does not utilize thought traces.
Results validate the effectiveness of thought traces for this task.
Additionally, as shown in Tab.~\ref{tab:ablation_study_benchmark}, we also observe that their benefits are reduced under SFT.

\textbf{Different Base LLMs and Tool Chain Lengths.}
We conduct experiments using different base LLMs in \workname.
Tab.~\ref{tab:baseLLM} shows that
Llama-3.1-8B-Instruct and Qwen2.5-7B-Instruct achieve comparable performance and outperform DeepSeek-R1-7B. 
We also test LLMs in 1.5B to 3B sizes.
More details are in Appendix~\ref{more_results}.
In addition, Tab.~\ref{tab:main_results_benchmark_level1}--Tab.~\ref{tab:main_results_benchmark_level3} show the performance of \workname~across samples with varying tool chain lengths.
We observe that most approaches achieve higher MD but lower FD, as the prompt we used encourages more conservative initiation of proactive services, leading to less intrusive assistance.
See Appendix~\ref{more_results} for more details.

\subsection{Out-of-Domain Evaluation}
\begin{figure}[h]
\vspace{-10pt}
    \centering
\includegraphics[width=\textwidth]{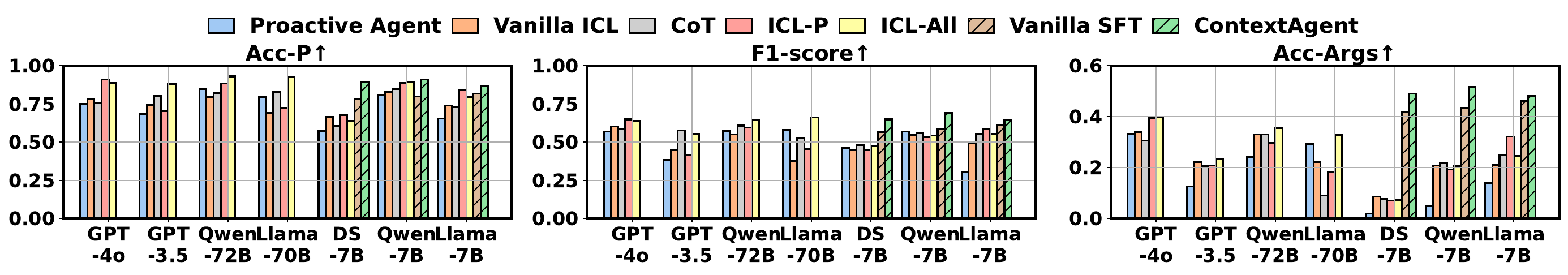}
\vspace{-15pt}
    \caption{Results on out-of-domain experiments.
    }
\vspace{-10pt}
    \label{fig:ood}
\end{figure}

We also evaluate \workname~under an out-of-distribution (OOD) setting. We randomly split \benchmark~based on scenarios. Samples from six scenarios are used for training, while those from the remaining three scenarios are used for evaluation.
Fig.~\ref{fig:ood} shows that \workname~achieves up to 90.9\% Acc-P, 68.9\% F1-score, and 51.6\% Acc-Args under OOD settings.
Furthermore, \workname~outperforms the best baseline by 1.9\% in Acc-Args, 10.7\% in F1-score, and 8.3\% in Acc-P, validating its generalization capabilities.
Tab.~\ref{tab:main_results_ood} in Appendix~\ref{full_results_on_benchmark} presents the complete results.

Besides, Tab.~\ref{tab:reasoning_llm} shows the performance of \workname~compared with proprietary and advanced reasoning LLMs under OOD settings. 
Results demonstrate that \workname~achieves comparable performance to these proprietary LLMs. 
Additionally, enhancing the reasoning capabilities of LLMs can further improve both the accuracy of proactive predictions and tool-calling performance.

\vspace{-10pt}

\begin{table}[h]
\setlength\tabcolsep{6pt}
\centering
\scriptsize
\caption{
Performance of \workname~compared with proprietary and advanced reasoning LLMs on \benchmark. \workname~employs DeepSeek-R1-7B as base model.}
\label{tab:reasoning_llm}
\begin{threeparttable}
\begin{tabular}{c|c|c|c|c|c|c|c|c}
\toprule
\multirow{2}{*}{\textbf{Settings}} &\multicolumn{4}{c|}{\textbf{Proactive Predictions}}  & \multicolumn{4}{c}{\textbf{Tool Calling}} \\ \cmidrule(l){2-9} 
 &  \textbf{Acc-P$^{\uparrow}$} &\textbf{MD.$^{\downarrow}$}
&\textbf{FD.$^{\downarrow}$}  & \textbf{RMSE$^{\downarrow}$}   & \textbf{Precision$^{\uparrow}$}  & \textbf{Recall$^{\uparrow}$}   & \textbf{F1-score$^{\uparrow}$}  & \textbf{Acc-Args$^{\uparrow}$}\\ 

\midrule
GPT-3.5-Turbo   & 0.879 & 0.020& 0.100 & 1.452 & 0.657& 0.521 &0.555 &0.235\\
GPT-4o   & 0.886 & 0.100& 0.013 & 1.701 & 0.719& 0.611 &0.639 &0.397\\
GPT-o4-mini   & 0.861 & 0.034& 0.103 & 1.240 & 0.726& 0.668 &0.682 &0.538\\
GPT-o3   & 0.868 & 0.069& 0.062 & 1.100 & 0.755& 0.697 &0.711 &0.563\\
Claude Sonnet 4  & 0.913 & 0.069& 0.017 & 1.010 & 0.775& 0.799 & 0.773& 0.480\\

\cellcolor{mygray}\textit{\workname}  & \cellcolor{mygray}0.893& \cellcolor{mygray}0.026&  \cellcolor{mygray}0.080& \cellcolor{mygray}1.249& \cellcolor{mygray}0.681& \cellcolor{mygray}0.645& \cellcolor{mygray}0.648& \cellcolor{mygray}0.489\\

\bottomrule
\end{tabular}
\end{threeparttable}
\vspace{-1em}
\end{table}
\section{Conclusion}
This paper introduces \workname, the first framework for context-aware proactive LLM agents.
\workname~can harness the context information from extensive sensory perceptions and tool-augmented LLM reasoning for enhanced proactive services.
To evaluate this new task, we further introduce \benchmark, the first benchmark for evaluating context-aware proactive LLM agents.
Our research takes a step towards further aligning with the vision of proactive AI assistants by leveraging rich context from hands-free wearable sensors to enhance proactive LLM reasoning.

\section*{Acknowledgments}
The research reported in this paper was partially supported by Research Grants Council of Hong Kong under grants 14207123, STG1/E-403/24-N, and National Science Foundation under Grant
Number CNS-1943396.
The views and conclusions contained here are those of the authors and should not be interpreted as necessarily representing the official policies or endorsements, either expressed or implied, of The
Chinese University of Hong Kong, Columbia University, NSF, or the U.S. Government or any of its agencies.

{\small
\bibliographystyle{abbrvnat}
\bibliography{ref}

\begin{thebibliography}{56}
\providecommand{\natexlab}[1]{#1}
\providecommand{\url}[1]{\texttt{#1}}
\expandafter\ifx\csname urlstyle\endcsname\relax
  \providecommand{\doi}[1]{doi: #1}\else
  \providecommand{\doi}{doi: \begingroup \urlstyle{rm}\Url}\fi

\bibitem[Pex(2024)]{Pexels}
Pexels.
\newblock \url{https://www.pexels.com/}, 2024.

\bibitem[Cla(2025)]{Claude-3.7-Sonnet}
Claude-3.7-sonnet.
\newblock \url{https://www.anthropic.com/claude/sonnet}, 2025.

\bibitem[fal(2025)]{fall_detection_apple_watch}
Use fall detection with apple watch.
\newblock \url{https://support.apple.com/en-hk/108896}, 2025.

\bibitem[Abbasian et~al.(2023)Abbasian, Azimi, Rahmani, and Jain]{abbasian2023conversational}
M.~Abbasian, I.~Azimi, A.~M. Rahmani, and R.~Jain.
\newblock Conversational health agents: A personalized llm-powered agent framework.
\newblock \emph{arXiv preprint arXiv:2310.02374}, 2023.

\bibitem[Abdelaziz et~al.(2024)Abdelaziz, Basu, Agarwal, Kumaravel, Stallone, Panda, Rizk, Bhargav, Crouse, Gunasekara, et~al.]{abdelaziz2024granite}
I.~Abdelaziz, K.~Basu, M.~Agarwal, S.~Kumaravel, M.~Stallone, R.~Panda, Y.~Rizk, G.~Bhargav, M.~Crouse, C.~Gunasekara, et~al.
\newblock Granite-function calling model: Introducing function calling abilities via multi-task learning of granular tasks.
\newblock \emph{arXiv preprint arXiv:2407.00121}, 2024.

\bibitem[Ahn et~al.(2022)Ahn, Brohan, Brown, Chebotar, Cortes, David, Finn, Fu, Gopalakrishnan, Hausman, et~al.]{ahn2022can}
M.~Ahn, A.~Brohan, N.~Brown, Y.~Chebotar, O.~Cortes, B.~David, C.~Finn, C.~Fu, K.~Gopalakrishnan, K.~Hausman, et~al.
\newblock Do as i can, not as i say: Grounding language in robotic affordances.
\newblock \emph{arXiv preprint arXiv:2204.01691}, 2022.

\bibitem[Bai et~al.(2025)Bai, Chen, Liu, Wang, Ge, Song, Dang, Wang, Wang, Tang, et~al.]{bai2025qwen2}
S.~Bai, K.~Chen, X.~Liu, J.~Wang, W.~Ge, S.~Song, K.~Dang, P.~Wang, S.~Wang, J.~Tang, et~al.
\newblock Qwen2. 5-vl technical report.
\newblock \emph{arXiv preprint arXiv:2502.13923}, 2025.

\bibitem[Basu et~al.(2024)Basu, Abdelaziz, Chaudhury, Dan, Crouse, Munawar, Kumaravel, Muthusamy, Kapanipathi, and Lastras]{basu2024api}
K.~Basu, I.~Abdelaziz, S.~Chaudhury, S.~Dan, M.~Crouse, A.~Munawar, S.~Kumaravel, V.~Muthusamy, P.~Kapanipathi, and L.~A. Lastras.
\newblock Api-blend: A comprehensive corpora for training and benchmarking api llms.
\newblock \emph{arXiv preprint arXiv:2402.15491}, 2024.

\bibitem[Cheng et~al.(2024)Cheng, Sun, Chu, Xu, Li, Zhang, and Wu]{cheng2024seeclick}
K.~Cheng, Q.~Sun, Y.~Chu, F.~Xu, Y.~Li, J.~Zhang, and Z.~Wu.
\newblock Seeclick: Harnessing gui grounding for advanced visual gui agents.
\newblock \emph{arXiv preprint arXiv:2401.10935}, 2024.

\bibitem[DeepSeek-AI(2025)]{deepseekai2025deepseekr1incentivizingreasoningcapability}
DeepSeek-AI.
\newblock Deepseek-r1: Incentivizing reasoning capability in llms via reinforcement learning, 2025.
\newblock URL \url{https://arxiv.org/abs/2501.12948}.

\bibitem[Deng et~al.(2024)Deng, Xu, Sun, Liu, Tan, Liu, Li, Luan, Wang, Yan, et~al.]{deng2024mobile}
S.~Deng, W.~Xu, H.~Sun, W.~Liu, T.~Tan, J.~Liu, A.~Li, J.~Luan, B.~Wang, R.~Yan, et~al.
\newblock Mobile-bench: An evaluation benchmark for llm-based mobile agents.
\newblock \emph{arXiv preprint arXiv:2407.00993}, 2024.

\bibitem[Deng et~al.(2023)Deng, Gu, Zheng, Chen, Stevens, Wang, Sun, and Su]{deng2023mind2web}
X.~Deng, Y.~Gu, B.~Zheng, S.~Chen, S.~Stevens, B.~Wang, H.~Sun, and Y.~Su.
\newblock Mind2web: Towards a generalist agent for the web.
\newblock \emph{Advances in Neural Information Processing Systems}, 36:\penalty0 28091--28114, 2023.

\bibitem[Englhardt et~al.(2024)Englhardt, Li, Nissanka, Zhang, Narayanswamy, Breda, Liu, Patel, and Iyer]{englhardt2024exploring}
Z.~Englhardt, R.~Li, D.~Nissanka, Z.~Zhang, G.~Narayanswamy, J.~Breda, X.~Liu, S.~Patel, and V.~Iyer.
\newblock Exploring and characterizing large language models for embedded system development and debugging.
\newblock In \emph{Extended Abstracts of the CHI Conference on Human Factors in Computing Systems}, pages 1--9, 2024.

\bibitem[Ge et~al.(2024)Ge, Chan, Wang, Yu, Mi, and Yu]{ge2024scaling}
T.~Ge, X.~Chan, X.~Wang, D.~Yu, H.~Mi, and D.~Yu.
\newblock Scaling synthetic data creation with 1,000,000,000 personas.
\newblock \emph{arXiv preprint arXiv:2406.20094}, 2024.

\bibitem[Grattafiori et~al.(2024)Grattafiori, Dubey, Jauhri, Pandey, Kadian, Al-Dahle, Letman, Mathur, Schelten, Vaughan, et~al.]{grattafiori2024llama}
A.~Grattafiori, A.~Dubey, A.~Jauhri, A.~Pandey, A.~Kadian, A.~Al-Dahle, A.~Letman, A.~Mathur, A.~Schelten, A.~Vaughan, et~al.
\newblock The llama 3 herd of models.
\newblock \emph{arXiv preprint arXiv:2407.21783}, 2024.

\bibitem[Guo et~al.(2025)Guo, Yang, Zhang, Song, Zhang, Xu, Zhu, Ma, Wang, Bi, et~al.]{guo2025deepseek}
D.~Guo, D.~Yang, H.~Zhang, J.~Song, R.~Zhang, R.~Xu, Q.~Zhu, S.~Ma, P.~Wang, X.~Bi, et~al.
\newblock Deepseek-r1: Incentivizing reasoning capability in llms via reinforcement learning.
\newblock \emph{arXiv preprint arXiv:2501.12948}, 2025.

\bibitem[Hurst et~al.(2024)Hurst, Lerer, Goucher, Perelman, Ramesh, Clark, Ostrow, Welihinda, Hayes, Radford, et~al.]{hurst2024gpt}
A.~Hurst, A.~Lerer, A.~P. Goucher, A.~Perelman, A.~Ramesh, A.~Clark, A.~Ostrow, A.~Welihinda, A.~Hayes, A.~Radford, et~al.
\newblock Gpt-4o system card.
\newblock \emph{arXiv preprint arXiv:2410.21276}, 2024.

\bibitem[Lee et~al.(2024)Lee, Choi, Lee, Wasi, Choi, Ko, Oh, and Shin]{lee2024mobilegpt}
S.~Lee, J.~Choi, J.~Lee, M.~H. Wasi, H.~Choi, S.~Ko, S.~Oh, and I.~Shin.
\newblock Mobilegpt: Augmenting llm with human-like app memory for mobile task automation.
\newblock In \emph{Proceedings of the 30th Annual International Conference on Mobile Computing and Networking}, pages 1119--1133, 2024.

\bibitem[Liu et~al.(2023)Liu, Li, Wu, and Lee]{liu2023visual}
H.~Liu, C.~Li, Q.~Wu, and Y.~J. Lee.
\newblock Visual instruction tuning.
\newblock \emph{Advances in neural information processing systems}, 36:\penalty0 34892--34916, 2023.

\bibitem[Liu et~al.(2025)Liu, Yang, Xu, Guo, Xing, Shuai, Ren, Jiang, and Yan]{liu2025tasksense}
K.~Liu, B.~Yang, L.~Xu, Y.~Guo, G.~Xing, X.~Shuai, X.~Ren, X.~Jiang, and Z.~Yan.
\newblock Tasksense: A translation-like approach for tasking heterogeneous sensor systems with llms.
\newblock In \emph{Proceedings of the 23rd ACM Conference on Embedded Networked Sensor Systems}, pages 213--225, 2025.

\bibitem[Liu et~al.(2024{\natexlab{a}})Liu, Huang, Zeng, Hao, Yu, Li, Wang, Gan, Liu, Yu, et~al.]{liu2024toolace}
W.~Liu, X.~Huang, X.~Zeng, X.~Hao, S.~Yu, D.~Li, S.~Wang, W.~Gan, Z.~Liu, Y.~Yu, et~al.
\newblock Toolace: Winning the points of llm function calling.
\newblock \emph{arXiv preprint arXiv:2409.00920}, 2024{\natexlab{a}}.

\bibitem[Liu et~al.(2024{\natexlab{b}})Liu, Wang, Wang, Xie, Qiu, Dang, Du, Chen, Yang, and Qian]{liu2024autonomous}
W.~Liu, C.~Wang, Y.~Wang, Z.~Xie, R.~Qiu, Y.~Dang, Z.~Du, W.~Chen, C.~Yang, and C.~Qian.
\newblock Autonomous agents for collaborative task under information asymmetry.
\newblock \emph{arXiv preprint arXiv:2406.14928}, 2024{\natexlab{b}}.

\bibitem[Liu et~al.(2024{\natexlab{c}})Liu, Fang, Shi, Wu, Igarashi, and Chen]{liu2024proactive}
X.~B. Liu, S.~Fang, W.~Shi, C.-S. Wu, T.~Igarashi, and X.~A. Chen.
\newblock Proactive conversational agents with inner thoughts.
\newblock \emph{arXiv preprint arXiv:2501.00383}, 2024{\natexlab{c}}.

\bibitem[Lu et~al.(2024)Lu, Yang, Qian, Chen, Luo, Wu, Wang, Cong, Zhang, Lin, et~al.]{lu2024proactive}
Y.~Lu, S.~Yang, C.~Qian, G.~Chen, Q.~Luo, Y.~Wu, H.~Wang, X.~Cong, Z.~Zhang, Y.~Lin, et~al.
\newblock Proactive agent: Shifting llm agents from reactive responses to active assistance.
\newblock \emph{arXiv preprint arXiv:2410.12361}, 2024.

\bibitem[Merrill et~al.(2024)Merrill, Paruchuri, Rezaei, Kovacs, Perez, Liu, Schenck, Hammerquist, Sunshine, Tailor, et~al.]{merrill2024transforming}
M.~A. Merrill, A.~Paruchuri, N.~Rezaei, G.~Kovacs, J.~Perez, Y.~Liu, E.~Schenck, N.~Hammerquist, J.~Sunshine, S.~Tailor, et~al.
\newblock Transforming wearable data into health insights using large language model agents.
\newblock \emph{arXiv preprint arXiv:2406.06464}, 2024.

\bibitem[Mialon et~al.(2023)Mialon, Fourrier, Wolf, LeCun, and Scialom]{mialon2023gaia}
G.~Mialon, C.~Fourrier, T.~Wolf, Y.~LeCun, and T.~Scialom.
\newblock Gaia: a benchmark for general ai assistants.
\newblock In \emph{The Twelfth International Conference on Learning Representations}, 2023.

\bibitem[Paranjape et~al.(2023)Paranjape, Lundberg, Singh, Hajishirzi, Zettlemoyer, and Ribeiro]{paranjape2023art}
B.~Paranjape, S.~Lundberg, S.~Singh, H.~Hajishirzi, L.~Zettlemoyer, and M.~T. Ribeiro.
\newblock Art: Automatic multi-step reasoning and tool-use for large language models.
\newblock \emph{arXiv preprint arXiv:2303.09014}, 2023.

\bibitem[Post et~al.(2025)Post, Kuchida, Olapade, Yin, Nurmi, and Flores]{post2025contextllm}
K.~Post, R.~Kuchida, M.~Olapade, Z.~Yin, P.~Nurmi, and H.~Flores.
\newblock Contextllm: Meaningful context reasoning from multi-sensor and multi-device data using llms.
\newblock In \emph{Proceedings of ACM HOTMOBILE'25}. Association for Computing Machinery (ACM), 2025.

\bibitem[Qin et~al.(2023)Qin, Liang, Ye, Zhu, Yan, Lu, Lin, Cong, Tang, Qian, et~al.]{qin2023toolllm}
Y.~Qin, S.~Liang, Y.~Ye, K.~Zhu, L.~Yan, Y.~Lu, Y.~Lin, X.~Cong, X.~Tang, B.~Qian, et~al.
\newblock Toolllm: Facilitating large language models to master 16000+ real-world apis.
\newblock \emph{arXiv preprint arXiv:2307.16789}, 2023.

\bibitem[Radford et~al.(2023)Radford, Kim, Xu, Brockman, McLeavey, and Sutskever]{radford2023robust}
A.~Radford, J.~W. Kim, T.~Xu, G.~Brockman, C.~McLeavey, and I.~Sutskever.
\newblock Robust speech recognition via large-scale weak supervision.
\newblock In \emph{International conference on machine learning}, pages 28492--28518. PMLR, 2023.

\bibitem[Shen et~al.(2025)Shen, Yang, Zheng, and Li]{shen2025autoiot}
L.~Shen, Q.~Yang, Y.~Zheng, and M.~Li.
\newblock Autoiot: Llm-driven automated natural language programming for aiot applications.
\newblock \emph{arXiv preprint arXiv:2503.05346}, 2025.

\bibitem[Shi et~al.(2024)Shi, Gao, Chen, Feng, Yan, Shi, Yin, Chen, Verberne, and Ren]{shi2024chain}
Z.~Shi, S.~Gao, X.~Chen, Y.~Feng, L.~Yan, H.~Shi, D.~Yin, Z.~Chen, S.~Verberne, and Z.~Ren.
\newblock Chain of tools: Large language model is an automatic multi-tool learner.
\newblock \emph{arXiv preprint arXiv:2405.16533}, 2024.

\bibitem[Shinn et~al.(2023)Shinn, Cassano, Gopinath, Narasimhan, and Yao]{shinn2023reflexion}
N.~Shinn, F.~Cassano, A.~Gopinath, K.~Narasimhan, and S.~Yao.
\newblock Reflexion: Language agents with verbal reinforcement learning.
\newblock \emph{Advances in Neural Information Processing Systems}, 36:\penalty0 8634--8652, 2023.

\bibitem[Trirat et~al.(2024)Trirat, Jeong, and Hwang]{trirat2024automl}
P.~Trirat, W.~Jeong, and S.~J. Hwang.
\newblock Automl-agent: A multi-agent llm framework for full-pipeline automl.
\newblock \emph{arXiv preprint arXiv:2410.02958}, 2024.

\bibitem[Wei et~al.(2022)Wei, Wang, Schuurmans, Bosma, Xia, Chi, Le, Zhou, et~al.]{wei2022chain}
J.~Wei, X.~Wang, D.~Schuurmans, M.~Bosma, F.~Xia, E.~Chi, Q.~V. Le, D.~Zhou, et~al.
\newblock Chain-of-thought prompting elicits reasoning in large language models.
\newblock \emph{Advances in neural information processing systems}, 35:\penalty0 24824--24837, 2022.

\bibitem[Wei et~al.(2025)Wei, Duchenne, Copet, Carbonneaux, Zhang, Fried, Synnaeve, Singh, and Wang]{wei2025swe}
Y.~Wei, O.~Duchenne, J.~Copet, Q.~Carbonneaux, L.~Zhang, D.~Fried, G.~Synnaeve, R.~Singh, and S.~I. Wang.
\newblock Swe-rl: Advancing llm reasoning via reinforcement learning on open software evolution.
\newblock \emph{arXiv preprint arXiv:2502.18449}, 2025.

\bibitem[Wen et~al.(2024)Wen, Li, Liu, Zhao, Yu, Li, Jiang, Liu, Zhang, and Liu]{wen2024autodroid}
H.~Wen, Y.~Li, G.~Liu, S.~Zhao, T.~Yu, T.~J.-J. Li, S.~Jiang, Y.~Liu, Y.~Zhang, and Y.~Liu.
\newblock Autodroid: Llm-powered task automation in android.
\newblock In \emph{Proceedings of the 30th Annual International Conference on Mobile Computing and Networking}, pages 543--557, 2024.

\bibitem[Wu et~al.(2024)Wu, Zhao, Huang, Huang, Yasunaga, Cao, Ioannidis, Subbian, Leskovec, and Zou]{wu2024avatar}
S.~Wu, S.~Zhao, Q.~Huang, K.~Huang, M.~Yasunaga, K.~Cao, V.~Ioannidis, K.~Subbian, J.~Leskovec, and J.~Y. Zou.
\newblock Avatar: Optimizing llm agents for tool usage via contrastive reasoning.
\newblock \emph{Advances in Neural Information Processing Systems}, 37:\penalty0 25981--26010, 2024.

\bibitem[Wu et~al.(2025)Wu, Wan, Li, Zhao, Ma, Ye, Pop, Zhang, and Chen]{wu2025proai}
Y.~Wu, G.~Wan, J.~Li, S.~Zhao, L.~Ma, T.~Ye, I.~Pop, Y.~Zhang, and J.~Chen.
\newblock Proai: Proactive multi-agent conversational ai with structured knowledge base for psychiatric diagnosis.
\newblock \emph{arXiv preprint arXiv:2502.20689}, 2025.

\bibitem[Yang et~al.(2024{\natexlab{a}})Yang, Yang, Zhang, Hui, Zheng, Yu, Li, Liu, Huang, Wei, et~al.]{yang2024qwen2}
A.~Yang, B.~Yang, B.~Zhang, B.~Hui, B.~Zheng, B.~Yu, C.~Li, D.~Liu, F.~Huang, H.~Wei, et~al.
\newblock Qwen2. 5 technical report.
\newblock \emph{arXiv preprint arXiv:2412.15115}, 2024{\natexlab{a}}.

\bibitem[Yang et~al.(2023)Yang, He, Ling, Yan, Xing, Shuai, Ren, and Jiang]{yang2023edgefm}
B.~Yang, L.~He, N.~Ling, Z.~Yan, G.~Xing, X.~Shuai, X.~Ren, and X.~Jiang.
\newblock Edgefm: Leveraging foundation model for open-set learning on the edge.
\newblock In \emph{Proceedings of the 21st ACM Conference on Embedded Networked Sensor Systems}, pages 111--124, 2023.

\bibitem[Yang et~al.(2024{\natexlab{b}})Yang, He, Liu, and Yan]{yang2024viassist}
B.~Yang, L.~He, K.~Liu, and Z.~Yan.
\newblock Viassist: Adapting multi-modal large language models for users with visual impairments.
\newblock In \emph{2024 IEEE International Workshop on Foundation Models for Cyber-Physical Systems \& Internet of Things (FMSys)}, pages 32--37. IEEE, 2024{\natexlab{b}}.

\bibitem[Yang et~al.(2024{\natexlab{c}})Yang, Jiang, Xu, Liu, Li, Xing, Chen, Jiang, and Yan]{yang2024drhouse}
B.~Yang, S.~Jiang, L.~Xu, K.~Liu, H.~Li, G.~Xing, H.~Chen, X.~Jiang, and Z.~Yan.
\newblock Drhouse: An llm-empowered diagnostic reasoning system through harnessing outcomes from sensor data and expert knowledge.
\newblock \emph{Proceedings of the ACM on Interactive, Mobile, Wearable and Ubiquitous Technologies}, 8\penalty0 (4):\penalty0 1--29, 2024{\natexlab{c}}.

\bibitem[Yang et~al.(2025)Yang, Guo, Xu, Yan, Chen, Xing, and Jiang]{yang2025socialmind}
B.~Yang, Y.~Guo, L.~Xu, Z.~Yan, H.~Chen, G.~Xing, and X.~Jiang.
\newblock Socialmind: Llm-based proactive ar social assistive system with human-like perception for in-situ live interactions.
\newblock \emph{Proceedings of the ACM on Interactive, Mobile, Wearable and Ubiquitous Technologies}, 9\penalty0 (1):\penalty0 1--30, 2025.

\bibitem[Yang et~al.(2024{\natexlab{d}})Yang, Li, Han, Li, and Xu]{yang2024embedgenius}
H.~Yang, M.~Li, M.~Han, Z.~Li, and W.~Xu.
\newblock Embedgenius: Towards automated software development for generic embedded iot systems.
\newblock \emph{arXiv preprint arXiv:2412.09058}, 2024{\natexlab{d}}.

\bibitem[Yang et~al.(2024{\natexlab{e}})Yang, Jimenez, Wettig, Lieret, Yao, Narasimhan, and Press]{yang2024swe}
J.~Yang, C.~Jimenez, A.~Wettig, K.~Lieret, S.~Yao, K.~Narasimhan, and O.~Press.
\newblock Swe-agent: Agent-computer interfaces enable automated software engineering.
\newblock \emph{Advances in Neural Information Processing Systems}, 37:\penalty0 50528--50652, 2024{\natexlab{e}}.

\bibitem[Yao et~al.(2023)Yao, Zhao, Yu, Du, Shafran, Narasimhan, and Cao]{yao2023react}
S.~Yao, J.~Zhao, D.~Yu, N.~Du, I.~Shafran, K.~Narasimhan, and Y.~Cao.
\newblock React: Synergizing reasoning and acting in language models.
\newblock In \emph{International Conference on Learning Representations (ICLR)}, 2023.

\bibitem[Ye et~al.(2023)Ye, Chen, Xu, Zu, Shao, Liu, Cui, Zhou, Gong, Shen, et~al.]{ye2023comprehensive}
J.~Ye, X.~Chen, N.~Xu, C.~Zu, Z.~Shao, S.~Liu, Y.~Cui, Z.~Zhou, C.~Gong, Y.~Shen, et~al.
\newblock A comprehensive capability analysis of gpt-3 and gpt-3.5 series models.
\newblock \emph{arXiv preprint arXiv:2303.10420}, 2023.

\bibitem[Zhang et~al.(2023)Zhang, Yang, Liu, Han, Chen, Huang, Fu, and Yu]{zhang2023appagent}
C.~Zhang, Z.~Yang, J.~Liu, Y.~Han, X.~Chen, Z.~Huang, B.~Fu, and G.~Yu.
\newblock Appagent: Multimodal agents as smartphone users.
\newblock \emph{arXiv preprint arXiv:2312.13771}, 2023.

\bibitem[Zhang et~al.(2024{\natexlab{a}})Zhang, Yang, Hu, Wang, Li, Sun, Zhang, Zhang, Liu, Zhu, et~al.]{zhang2024proagent}
C.~Zhang, K.~Yang, S.~Hu, Z.~Wang, G.~Li, Y.~Sun, C.~Zhang, Z.~Zhang, A.~Liu, S.-C. Zhu, et~al.
\newblock Proagent: building proactive cooperative agents with large language models.
\newblock In \emph{Proceedings of the AAAI Conference on Artificial Intelligence}, volume~38, pages 17591--17599, 2024{\natexlab{a}}.

\bibitem[Zhang et~al.(2024{\natexlab{b}})Zhang, Li, Li, Shi, and Jin]{zhang2024codeagent}
K.~Zhang, J.~Li, G.~Li, X.~Shi, and Z.~Jin.
\newblock Codeagent: Enhancing code generation with tool-integrated agent systems for real-world repo-level coding challenges.
\newblock \emph{arXiv preprint arXiv:2401.07339}, 2024{\natexlab{b}}.

\bibitem[Zhang et~al.(2024{\natexlab{c}})Zhang, Deng, Ren, Ng, and Chua]{zhang2024ask}
X.~Zhang, Y.~Deng, Z.~Ren, S.-K. Ng, and T.-S. Chua.
\newblock Ask-before-plan: Proactive language agents for real-world planning.
\newblock \emph{arXiv preprint arXiv:2406.12639}, 2024{\natexlab{c}}.

\bibitem[Zhang et~al.(2024{\natexlab{d}})Zhang, Sun, Chen, Pfister, Zhang, and Arik]{zhang2024chain}
Y.~Zhang, R.~Sun, Y.~Chen, T.~Pfister, R.~Zhang, and S.~Arik.
\newblock Chain of agents: Large language models collaborating on long-context tasks.
\newblock \emph{Advances in Neural Information Processing Systems}, 37:\penalty0 132208--132237, 2024{\natexlab{d}}.

\bibitem[Zhao et~al.(2024)Zhao, Huang, Xu, Lin, Liu, and Huang]{zhao2024expel}
A.~Zhao, D.~Huang, Q.~Xu, M.~Lin, Y.-J. Liu, and G.~Huang.
\newblock Expel: Llm agents are experiential learners.
\newblock In \emph{Proceedings of the AAAI Conference on Artificial Intelligence}, volume~38, pages 19632--19642, 2024.

\bibitem[Zhao et~al.(2025)Zhao, Zhu, Mozannar, Sontag, Talwalkar, and Chen]{zhao2025codinggenie}
S.~Zhao, A.~Zhu, H.~Mozannar, D.~Sontag, A.~Talwalkar, and V.~Chen.
\newblock Codinggenie: A proactive llm-powered programming assistant.
\newblock \emph{arXiv preprint arXiv:2503.14724}, 2025.

\bibitem[Zheng et~al.(2024)Zheng, Gou, Kil, Sun, and Su]{zheng2024gpt}
B.~Zheng, B.~Gou, J.~Kil, H.~Sun, and Y.~Su.
\newblock Gpt-4v (ision) is a generalist web agent, if grounded.
\newblock \emph{arXiv preprint arXiv:2401.01614}, 2024.

\end{thebibliography}
}


\clearpage
\appendix
\setcounter{footnote}{0}
\newpage
\section*{Appendix}

\section{Data Examples}
We first provide several examples in our dataset, including cases with different proactive scores.

\noindent\makebox[\linewidth]{\rule{\linewidth}{0.4pt}}\\
\textit{Example 1}
\begin{lstlisting}[basicstyle=\ttfamily, breaklines=true]
{
    "Context information": "Visual information suggests that the user is conversing with a partner on the road. Audio information shows there is a conversation between the user and others. The partner says "Hi, how about organizing a team-building hike this weekend? We haven't had a team-building event in a long time."",
    
    "Personas": [
        "A software engineer specializing in document management systems, working closely with the graphic designer to implement the designed interfaces."
    ],
    
    "Thoughts": "When planning hiking activities, various factors should be considered, including the weather, whether it is the off-season or peak season for tourism, passenger flow, schedule, and more. Since the hiking location has not been specified, it may be in the city where the user is currently located. I need to use GPS to determine the user's location.",
    
    "Proactive score": 5,
    
    "Tools": "[{"name": "get_current_gps_coordinates", "desc": "Get the current GPS coordinates of the user", "params": "None"}, {"name": "get_city_weather", "desc": "Get the weather for a specified city at a given time.", "params": {"city": "$RESULT(get_current_gps_coordinates.city)", "time": "this weekend"}}, {"name": "get_current_datetime", "desc": "Get the current date and time", "params": "None"}, {"name": "check_agenda_time_conflict", "desc": "Check if there is a time conflict in the user's agenda for a given datetime.", "params": "None"}]",

    "Response": "The weather for the hiking location this weekend looks perfect-clear skies and mild temperatures. Do you want me to help you pack or suggest what to bring?"
}
\end{lstlisting}

\noindent\makebox[\linewidth]{\rule{\linewidth}{0.4pt}}\\
\textit{Example 2}
\begin{lstlisting}[basicstyle=\ttfamily, breaklines=true]
{
    "Context information": "Visual information shows the user is sitting at a dining table with a plate of food that includes pasta, grilled chicken, and a side of salad.",
    
    "Personas": [
        "The user has a history of managing Type 2 diabetes.",
        "The user is health-conscious and actively manages their diet to control blood sugar levels."
    ],
    
    "Thoughts": "The user's personas show his glucose level is slightly elevated, and the user might appreciate dietary advice to help stabilize it. Providing this information could be helpful without being intrusive.",
    
    "Proactive score": 4,
    
    "Tools": "[{"name": "get_health_data", "desc": "Get health data from the user's smart device.", "params": "None"},{"name": "wikipedia_search", "desc": "Tool that searches the Wikipedia API. Useful for when you need to answer general questions about people, places, companies, facts, historical events, or other subjects.", "params": {"query": "The effect of pasta, grilled chicken, and a side of salad on stabilizing blood sugar levels."}}]",

    "Response": "Your glucose level appears slightly elevated. To help stabilize it, you might consider increasing the portion of salad or reducing the amount of pasta. Grilled chicken is a great choice for protein, as it has minimal impact on blood sugar levels. Would you like me to provide additional dietary tips or help track your meal for better management?"
}
\end{lstlisting}

\noindent\makebox[\linewidth]{\rule{\linewidth}{0.4pt}}\\
\textit{Example 3}
\begin{lstlisting}[basicstyle=\ttfamily, breaklines=true]
{
    "Context information": "Visual information suggests that the user is looking in the mirror while wearing summer clothes, such as short sleeves and shorts.",
    
    "Personas": [
        "A university psychology student who is currently studying creativity and personality."
    ],
    
    "Thoughts": "The user is looking in the mirror and may be preparing to go out, wearing minimal clothing such as short sleeves and shorts. I need to check the weather for the user's location. If the temperature is expected to drop or if rain is forecasted for the day, I can proactively remind the user to dress warmly to avoid catching a cold.",
    
    "Proactive score": 3,
    
    "Tools": "[{"name": "get_current_gps_coordinates", "desc": "Get the current GPS coordinates of the user", "params": "None"},{"name": "get_city_weather", "desc": "Get the weather for a specified city at a given time.", "params": {"city": "$RESULT(get_current_gps_coordinates.city)", "time":"now"}}]",

    "Response": "I checked the weather forecast, and the local temperature is expected to drop sharply after 6 PM. Considering that you are only wearing short-sleeved shirts and shorts, it is recommended that you add more layers."
}
\end{lstlisting}

\noindent\makebox[\linewidth]{\rule{\linewidth}{0.4pt}}\\
\textit{Example 4}
\begin{lstlisting}[basicstyle=\ttfamily, breaklines=true]
{
    "Context information": "Visual information suggests the user is in a gym, lifting weights.",
    
    "Personas": [
        "A computer science major interested in developing software for audio manipulation and enhancement."
    ],
    
    "Thoughts": "The user might need guidance on proper weightlifting techniques or reminders to take breaks between sets. However, without detailed persona information, such as their fitness level, I would rate this proactive behavior at 2 points. This rating indicates some potential for proactivity, but the necessity is not very high. After collecting more detailed persona information, I could provide more accurate proactive services.",
    
    "Proactive score": 2,
    "Tools": "None",
    "Response": "None"
}
\end{lstlisting}

\noindent\makebox[\linewidth]{\rule{\linewidth}{0.4pt}}\\
\textit{Example 5}
\begin{lstlisting}[basicstyle=\ttfamily, breaklines=true]
{
    "Context information": "Visual information shows the user is walking on the road. There are many trees on both sides of the road, and the road seems to be going uphill.",
    
    "Personas": [
        "A dog owner who wants the best medical treatment for their furry friend."
    ],
    
    "Thoughts": "Contextual information shows user is walking on the road. There are many trees on both sides of the road, and the road seems to be going uphill. Without explicit input or additional contextual information, no proactive actions are required. The proactive score is set to 1.",
    
    "Proactive score": 1,
    "Tools": "None",
    "Response": "None"
}
\end{lstlisting}

\section{Prompts}
\label{sec_prompts}
This section introduces the details of the prompts used in this work, including the system prompt used in \workname~and baselines, the prompt used in the data generation pipeline, and the prompt used in proactive-oriented context extraction.

\begin{figure}[htb]
  \centering
\includegraphics[width=1\linewidth]{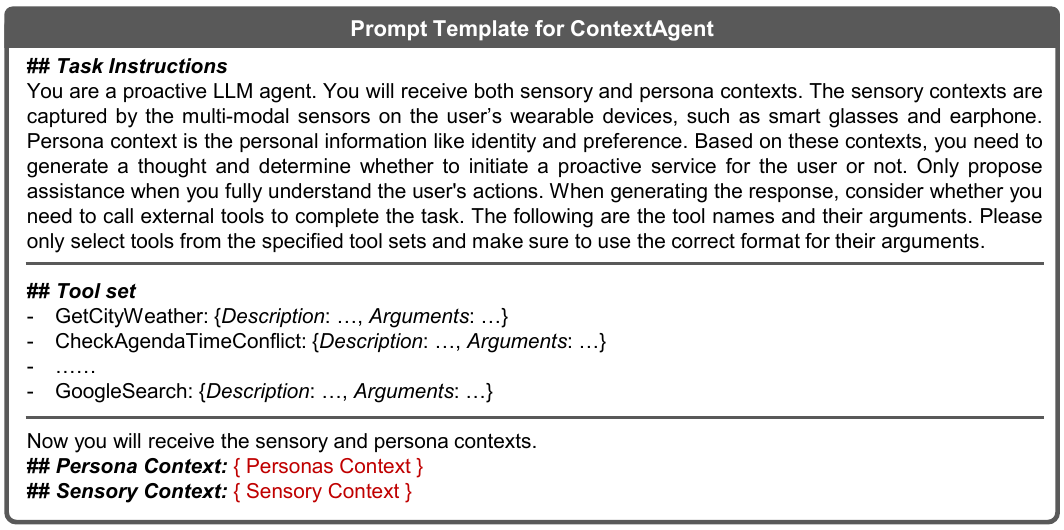}
  \caption{System prompt for \workname.}
\label{fig:prompt-contextagent}
\end{figure}

\textbf{System Prompt in \workname.}
Fig.~\ref{fig:prompt-contextagent} shows the prompt used in \workname.
It contains both the static prompt and the runtime prompt.
The static prompt includes task instructions and toolset definitions that remain constant throughout. The task instructions guide the LLM in understanding its role as a context-aware proactive agent and highlight the key considerations for this task.
The toolset definitions allow LLM agents to identify the available external tools and understand how to use them, including their names, arguments, and formats (See Appendix~\ref{sec:tool_def}).
The runtime prompt includes user personas and contextual information, which vary across different samples in the dataset.

\begin{figure}[htb]
  \centering
\includegraphics[width=1\linewidth]{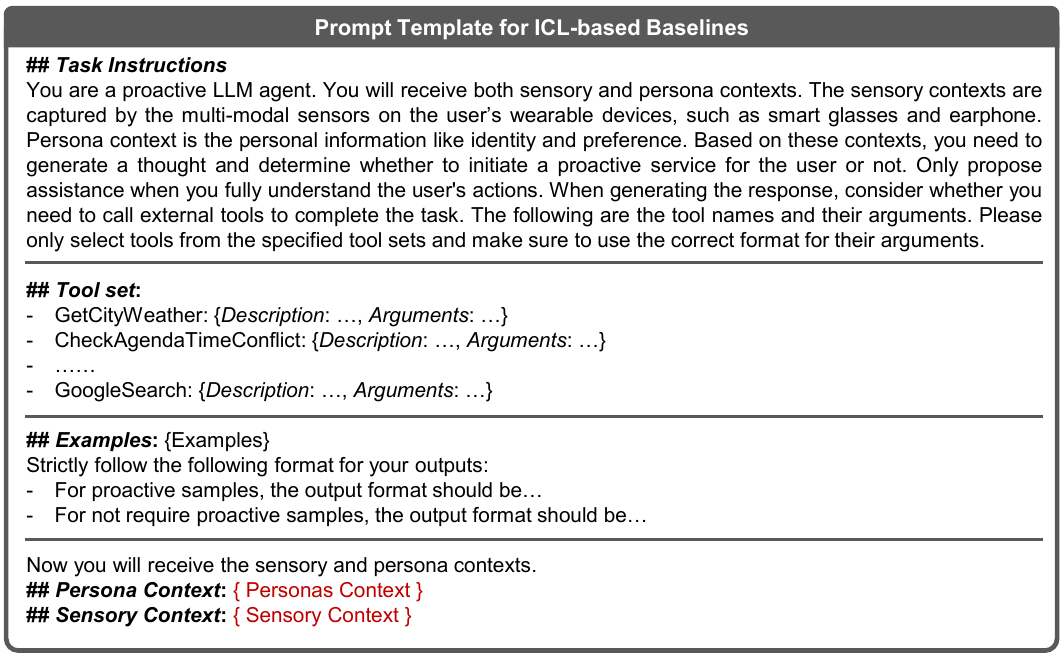}
  \caption{Prompt template for baselines.}
\label{fig:prompt-baselines}
  \vspace{-1.em}
\end{figure}

\begin{figure}[t]
  \centering
\includegraphics[width=1\linewidth]{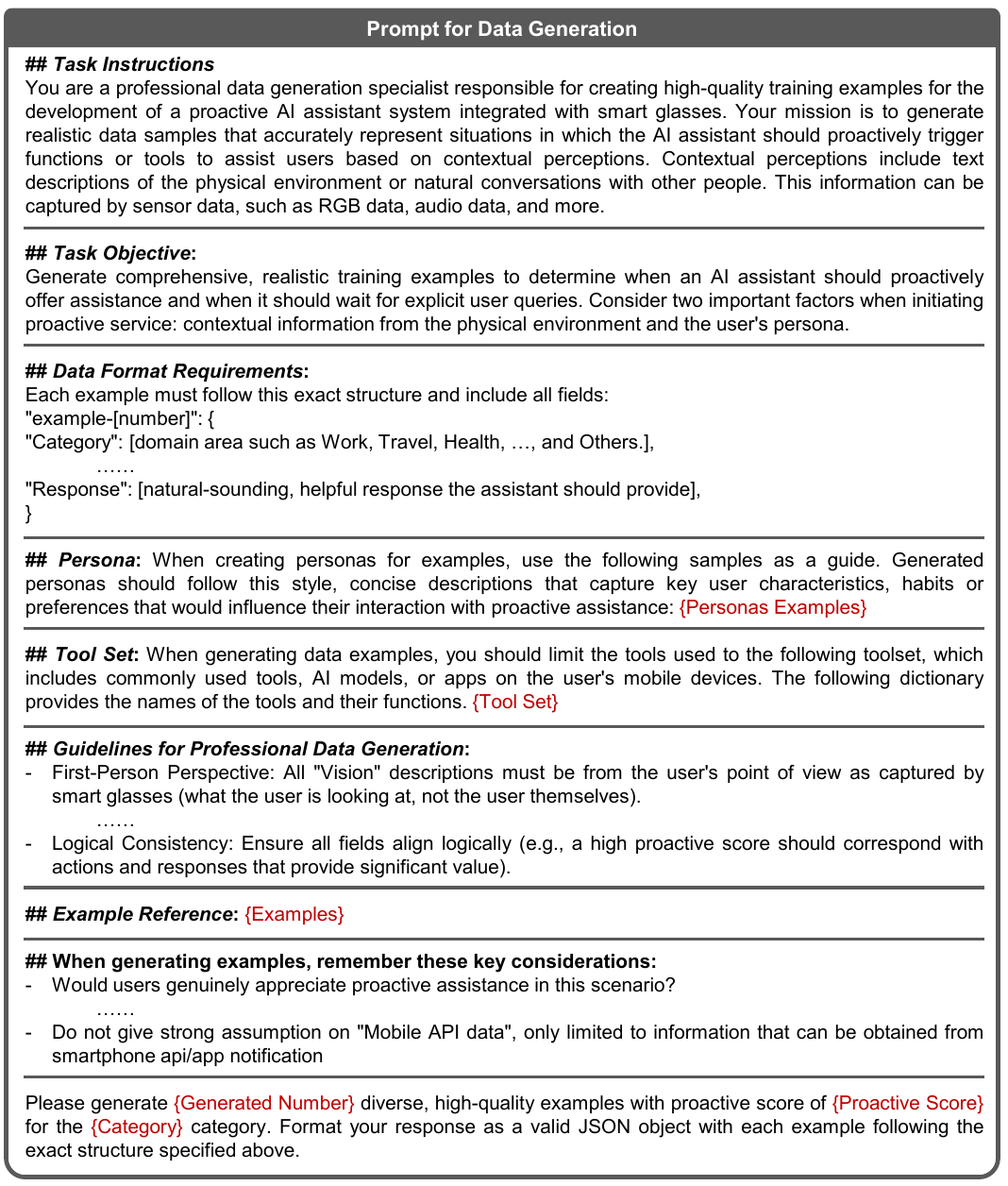}
  \caption{Prompt template for the data generation pipeline.}
\label{fig:prompt-data-generation}
  \vspace{-1.em}
\end{figure}

\begin{figure}[t]
  \centering
\includegraphics[width=1\linewidth]{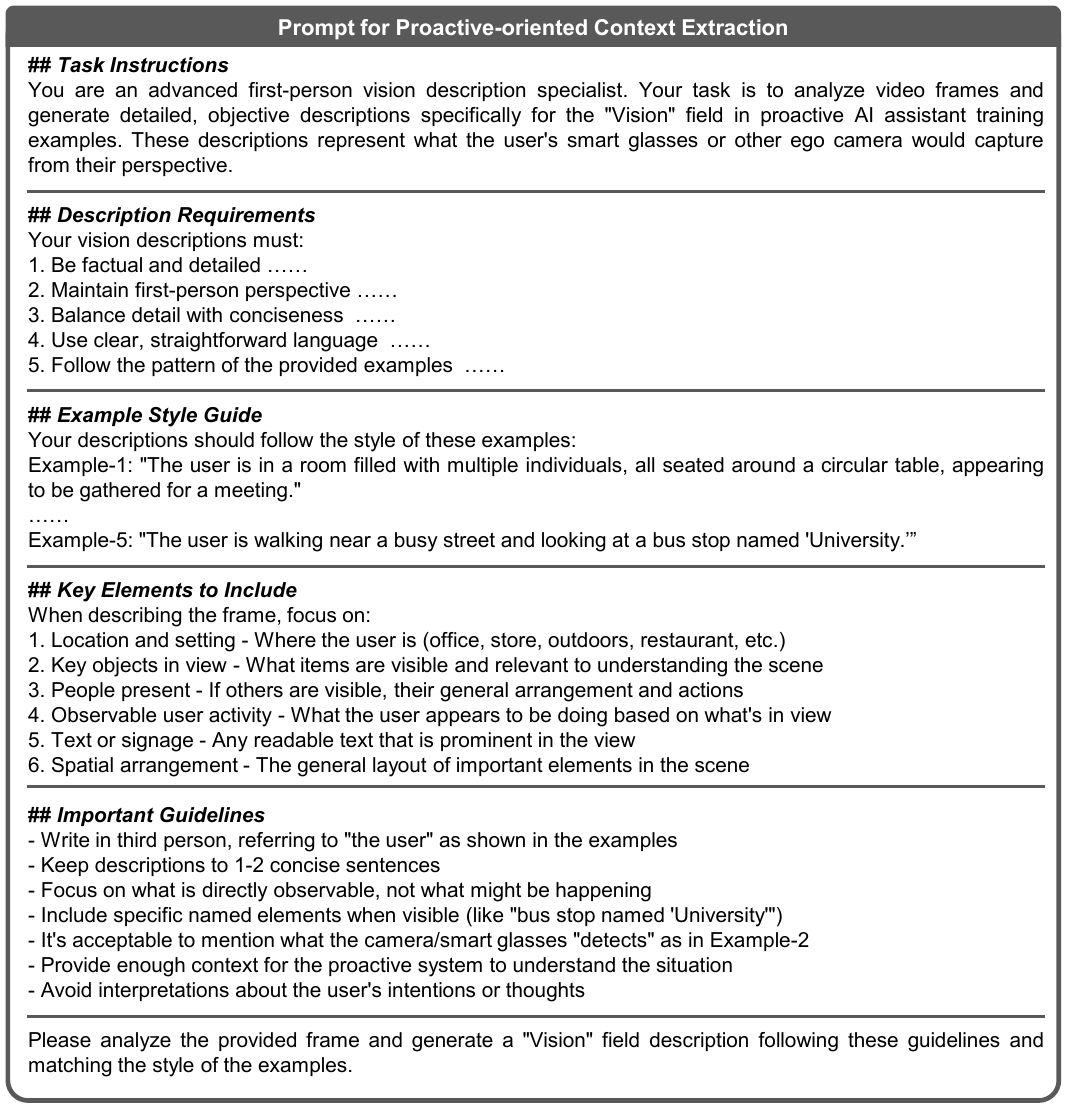}
  \caption{Prompt template for the proactive-oriented context extraction.}
\label{fig:prompt-proactive-context-extraction}
  \vspace{-1.em}
\end{figure}

\textbf{System Prompt for Baselines.}
Fig.~\ref{fig:prompt-baselines} shows the prompt template used in baseline approaches.
To ensure a fair comparison, we keep the task instructions and toolset definitions in the prompt the same as those used in \workname.
For the baseline of Proactive Agent, we do not include any samples in the example part of the prompt template.
For Vanilla ICL, CoT, ICL-P, and ICL-All, we randomly select ten samples from the training set and incorporate them into the prompt.
In Vanilla ICL, the few-shot examples contain only sensory context. In CoT, we additionally include thought traces in the few-shot examples. For ICL-P, we incorporate persona context into the few-shot demonstrations, and for ICL-All, we integrate sensory context, persona context, and thought traces into the few-shot demonstrations.
Additionally, for the Vanilla SFT baseline, we use the prompt template shown in Fig.~\ref{fig:prompt-contextagent}, but without including the persona context.

\textbf{Prompt for Data Generation Pipeline.} Fig.~\ref{fig:prompt-data-generation} shows the prompt template used for data generation.
We include the required format of data, examples of persona\footnote{\url{https://github.com/tencent-ailab/persona-hub}} and completed data samples, and detailed descriptions of the accessible tools. Besides, it specifies the structure and formulation of each component within a data sample to guide LLMs to generate high-quality samples.
During data generation, we use two strategies, including scenario-aware and proactive score-aware.
LLMs are guided to generate samples for different scenarios and target proactive scores separately, which enables LLMs to more effectively learn scenario-specific and score-specific patterns from the provided examples.
And when generating samples requiring proactive services, LLMs are instructed to consider using tools from our tool set. 
To further enhance the diversity of the generated samples, we employ various LLMs, including GPT-4o\footnote{\url{https://openai.com/index/gpt-4o-system-card/}}, Claude-3.7-Sonnet\footnote{\url{https://www.anthropic.com/claude/sonnet}}, and Gemini-2.0-Flash\footnote{\url{https://deepmind.google/technologies/gemini/flash/}}, for generation.

\begin{wrapfigure}{r}{0.58\textwidth}
\vspace*{-5mm}
\begin{center}
\includegraphics[width=\linewidth]{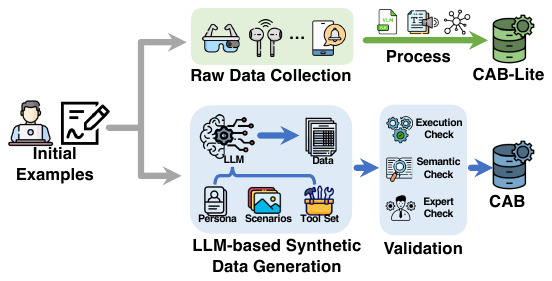}
\end{center}
\vspace{-10pt}
\caption{Flowchart illustrating the pipeline for constructing our benchmark. 
`CAB' refers to `ContextAgentBech'.
}
\label{fig:dataset_pipeline}
\vspace{-10pt}
\end{wrapfigure}

\textbf{Prompt for Proactive-oriented Context Extraction.}
Fig.~\ref{fig:prompt-proactive-context-extraction} shows the prompt template used for proactive-oriented context extraction.
With this prompt, the VLM focuses on objective, detailed scene understanding and key cues capturing (e.g., location, objects, and user actions) that are critical for determining whether proactive assistance is needed.
To ensure consistency and output quality, we include five example descriptions within the prompt. These examples illustrate the expected level of detail, structure, and tone, enabling the model to align its output with the expected format.
Based on the clear content and formatting guidelines, generated descriptions can provide high-quality context information for the following tasks, including predicting the necessity of proactive services and tool planning.
In this study, we employ Qwen-2.5-VL~\cite{bai2025qwen2} and Whisper~\cite{radford2023robust} to extract the visual and audio contexts, respectively.

\section{Dataset Details}
\label{dataset_details}
As described in Sec.~\ref{sec:dataset_construction}, the annotators first brainstorm and design initial exemplars to construct the seed dataset.
We then utilize an automated diversification pipeline to scale the dataset. 
Fig.~\ref{fig:dataset_pipeline} shows the flowchart of our pipeline to generate the~\benchmark~(CAB) and~\benchmarklite~(CAB-Lite). 
Additionally, we also collect raw video and audio data based on the textual descriptions of sensory contexts in the dataset.
This raw data is sourced from both copyright-free internet platforms\footnote{\url{https://www.pexels.com/license/}} and our own collections.
The self-collected samples are primarily used to gather chitchat scenarios.
This study has received IRB approval. All participants offered informed consent before any data were collected.
Additionally, we have blurred sensitive regions in the video, such as faces, to ensure participants' privacy.

\begin{table}[htbp]
\renewcommand{\arraystretch}{1.5}
  \caption{The definition of the 20 tools used in \benchmark.}
  \label{tab:tool_desc}
  \centering
  \resizebox{0.99\textwidth}{!}{
  \begin{tabular}{l|p{4.5cm}|p{4.5cm}|p{4.5cm}|p{4.5cm}}
    \toprule
\textbf{Index}  &\textbf{Name}  &\textbf{Description} & \textbf{Input} & \textbf{Output} \\
 \midrule

1 & GetCityWeather&Get the weather for a specified city at a given time. &[text] The city to fetch weather for. [text] The time to fetch weather for. &[text] Weather condition for a specified city at a given time. \\
 \midrule
2 & DateTime &Get the current date and time. &None. &[text] Current date and time. \\
\midrule
3 & CheckAgendaTimeConflict &Check if there is a time conflict in the user's agenda for a given datetime
and return all events as a summarized string.&[text] The time to check for conflicts. &[text] A summary of all events and whether there is a conflict. \\
\midrule
4 & WikipediaSearch & Search on Wikipedia.&[text] Search query. &[text] Wikipedia search result. \\
\midrule
5 & GetCurrentGPS. &Get the current GPS coordinates of the user.&None. &[text] GPS coordinates of the user. \\
 \midrule
6 & GetOnlineProductPrice &Get the price of a product from an online store.&[text] The name of the product to search for. &[text] The price of the product as a string. \\
\midrule
7 &SearchRednote &A platform where people share tips on travel, fitness, cooking, and more, allowing users to search for relevant strategies. &[text] The search query. &[text] The search results from rednote. \\
 \midrule
8 & VisualLanguageModel &Visual Language Model that can answer the user's questions based on the given image.& [image] Any image. [text] The prompt containing the user's question.& [text] The response from the VLLM.\\
\midrule
9 & GoogleMap &Get the route and distance from the current location to the destination using Google Maps API.&[text] The starting location. [text] The destination location. & [text] The route and distance information.\\
\midrule
10 &BookUber &Book an Uber ride from the current location to the destination.&[text] The starting location. [text] The destination location.&[text] The Uber ride booking confirmation.  \\
\midrule
11 & GetHealthData &Get health data from the user's smart device.& None. & [text] The health data as a string. \\
 \midrule
12 & GetMedicalKnowledge &Get medical expert knowledge from the up-to-date medical knowledge database.&[text] The query string containing the medical topic or symptoms. & [text] The medical expert knowledge as a string. \\
\midrule
13 & PlayMusic &Play a song from the user's music library.&None. &[text] The song playing confirmation. \\
\midrule
14 & AddtoAgenda &Add an event to the user's agenda.&[text] The name of the event to add. [text] The time of the event. &[text] The confirmation message. \\
 \midrule
15 &CheckBusSchedule &Check the bus schedule for a specific bus stop.&[text] The name of the bus stop.&[text] The bus schedule information.  \\
\midrule
16 &GoogleSearch &Search on Google. & [text] Search query. &[text] Description of the search result. \\
\midrule
17 &SetTimer &Set a timer for a specific duration.&[text] The duration of the timer. &[text] The timer set confirmation. \\
 \midrule
18 &QueryStock &This API queries the stock price of a given stock code and date.&[text] The stock code of the given stock. [text] The date of the stock price. &[text] The stock price of the given stock. \\
\midrule
19 &AddMeeting &This API allows users to make a reservation for a meeting and store the meeting information (e.g., topic, time, location, attendees) in the database.&[text] The topic of the meeting. [text] The start time of the meeting. [text] The location where the meeting to be held. &[text] Success or failed. \\
\midrule
20 &SendEmail &This API for sending email, given the receiver, subject and content.&[text] The receiver address of the email. [text] The subject address of the email. [text] The content of the email. &[text] The status of the email. \\
\bottomrule
  \end{tabular}
}
\end{table}

\section{Tool Definition}
\label{sec:tool_def}
\benchmark~contains 20 tools.
Given that the main focus of this study is to bridge the sensory context with tool-based LLM agents, we adopt the definitions of tools from existing work\footnote{\url{https://github.com/AlibabaResearch/DAMO-ConvAI/tree/main/api-bank/}}.
Tab.~\ref{tab:tool_desc} provides detailed information about the tools, including their names, tool descriptions, and input and output arguments.

\begin{table}[t]
\setlength\tabcolsep{5pt}
\centering
\scriptsize
\caption{
Main results on \benchmark.
}
\label{tab:main_results_benchmark_all}
\begin{threeparttable}
\begin{tabular}{c|c|c|c|c|c|c|c|c|c}
\toprule
\multirow{2}{*}{\textbf{Model}}     & \multirow{2}{*}{\textbf{Method}} &\multicolumn{4}{c|}{\textbf{Proactive Predictions}}  & \multicolumn{4}{c}{\textbf{Tool Calling}}  \\ \cmidrule(l){3-10} 
&  & \textbf{Acc-P$^{\uparrow}$} &\textbf{MD.$^{\downarrow}$}
&\textbf{FD.$^{\downarrow}$}  & \textbf{RMSE$^{\downarrow}$}   & \textbf{Precision$^{\uparrow}$}  & \textbf{Recall$^{\uparrow}$}   & \textbf{F1-score$^{\uparrow}$}  & \textbf{Acc-Args$^{\uparrow}$}  \\

\midrule
\multirow{5}{*}{GPT-4o}    & Proactive Agent& 0.785& 0.197& 0.017 &2.239 & 0.696 & 0.591 & 0.618& 0.246\\
& Vanilla ICL & 0.816& 0.166& 0.017 & 2.079& 0.703 & 0.614 &0.635 & 0.462\\
& CoT  & 0.833& 0.142& 0.023 &2.016 & 0.732 &0.645  &0.666 &0.529 \\
& ICL-P  & 0.921& 0.061& 0.017 &1.507 & 0.797 & 0.685 &0.714 &0.596 \\
& ICL-All  & 0.921& 0.057& 0.020 & 1.472& 0.788 & 0.690 & 0.715& 0.596\\

\midrule
\multirow{5}{*}{GPT-3.5-Turbo}    & Proactive Agent& 0.673& 0.010& 0.316 & 1.815& 0.493 & 0.380 & 0.410& 0.065 \\
& Vanilla ICL   & 0.772& 0.064& 0.163 & 1.710& 0.602 & 0.490 & 0.518& 0.374\\
& CoT  & 0.836 &0.088 & 0.074 & 1.534& 0.662 & 0.584 & 0.604& 0.433\\
& ICL-P  & 0.806& 0.040& 0.153 &1.713 & 0.634 & 0.517 &0.547 &0.400 \\
& ICL-All  & 0.846& 0.054& 0.098 & 1.439& 0.681 & 0.595 & 0.616& 0.458\\

\midrule
\multirow{5}{*}{Qwen2.5-72B-Ins}    & Proactive Agent& 0.843& 0.064& 0.091 & 1.717& 0.670 & 0.555 & 0.585& 0.144\\
& Vanilla ICL   & 0.867& 0.078& 0.054 & 1.324& 0.695 & 0.625 & 0.642& 0.475\\
& CoT  & 0.884& 0.071& 0.044 & 1.254& 0.708 & 0.643 & 0.660& 0.468\\
& ICL-P  & 0.915& 0.040& 0.044 &1.078 & 0.774 & 0.687 & 0.711& 0.495\\
& ICL-All  & 0.935& 0.037& 0.027 & 1.059& 0.750 & 0.688 & 0.703& 0.474\\

\midrule
\multirow{5}{*}{Llama3.1-70B-Ins}   & Proactive Agent& 0.850& 0.061& 0.088 & 1.643& 0.642 & 0.554 & 0.578& 0.314\\
& Vanilla ICL   & 0.799& 0.013& 0.187 & 1.442& 0.582 & 0.501 & 0.524& 0.374\\
& CoT  & 0.850& 0.020& 0.129 & 1.237& 0.637 & 0.543 &0.566 &0.385 \\
& ICL-P  & 0.857& 0.000& 0.143 & 1.322& 0.660 & 0.578 & 0.599& 0.375\\
& ICL-All  & 0.908& 0.003& 0.088 & 1.061& 0.712 & 0.644 & 0.660& 0.455 \\

\midrule
\multirow{7}{*}{DeepSeek-R1-7B}     & Proactive Agent& 0.544& 0.411& 0.044 & 3.093& 0.467 & 0.454 & 0.457&0.019 \\
& Vanilla ICL   & 0.646& 0.248& 0.105 & 2.568& 0.457 & 0.433 & 0.437& 0.078\\
& CoT  & 0.653& 0.319& 0.027 & 2.760& 0.528 & 0.501 & 0.507& 0.109\\
& ICL-P  & 0.690&  0.227 & 0.081& 2.466 & 0.518& 0.479& 0.486& 0.126\\
& ICL-All  & 0.704& 0.268& 0.027 & 2.540& 0.545 & 0.510 & 0.518& 0.103\\
& Vanilla SFT  & 0.823& 0.068&  0.108& 1.630& 0.621&  0.570& 0.579& 0.386\\
&\cellcolor{mygray}\textit{\workname}  & \cellcolor{mygray}0.888& \cellcolor{mygray}0.027&  \cellcolor{mygray}0.085& \cellcolor{mygray}1.319& \cellcolor{mygray}0.676& \cellcolor{mygray}0.648& \cellcolor{mygray}0.647& \cellcolor{mygray}0.468\\

\midrule
\multirow{7}{*}{Qwen2.5-7B-Ins}     & Proactive Agent& 0.799& 0.136& 0.064 & 2.038& 0.578 & 0.536 & 0.546& 0.038\\
& Vanilla ICL   & 0.816& 0.088& 0.095 & 1.752& 0.590 & 0.545 & 0.553& 0.326\\
& CoT  & 0.833& 0.085& 0.081 & 1.790& 0.585 & 0.527 & 0.541& 0.272\\
& ICL-P  & 0.833& 0.091& 0.074 & 1.819& 0.610 & 0.556 & 0.568& 0.303\\
& ICL-All  &0.867 & 0.088& 0.044 & 1.721& 0.635 & 0.577 & 0.591& 0.301\\
& Vanilla SFT  & 0.775& 0.088& 0.136 & 1.774& 0.589 & 0.551 & 0.558&0.398\\
& \cellcolor{mygray}\textit{\workname}  &\cellcolor{mygray}0.894 &\cellcolor{mygray}0.013 & \cellcolor{mygray}0.091 & \cellcolor{mygray}1.264& \cellcolor{mygray}0.672 & \cellcolor{mygray}0.644 & \cellcolor{mygray}0.645& \cellcolor{mygray}0.459\\

\midrule
\multirow{7}{*}{Llama3.1-8B-Ins}     & Proactive Agent& 0.690& 0.006& 0.302 & 1.831 & 0.376 &0.280 &0.305 & 0.091 \\
& Vanilla ICL   & 0.748& 0.193& 0.057 & 1.898& 0.612 & 0.526 & 0.548& 0.234\\
& CoT  & 0.768& 0.159& 0.071 & 1.770& 0.596 &0.512  & 0.533& 0.257\\
& ICL-P  & 0.724& 0.268& 0.006 &2.207 &0.658  & 0.563 & 0.587& 0.251\\
& ICL-All  & 0.833& 0.139& 0.027 &1.624 & 0.662 & 0.573 & 0.596& 0.298\\
& Vanilla SFT  & 0.819& 0.071& 0.108 & 1.650& 0.597 & 0.567 & 0.567&0.362\\
&\cellcolor{mygray}\textit{\workname}  & \cellcolor{mygray}0.874& \cellcolor{mygray}0.030&  \cellcolor{mygray}0.095& \cellcolor{mygray}1.408& \cellcolor{mygray}0.660& \cellcolor{mygray}0.627& \cellcolor{mygray}0.626& \cellcolor{mygray}0.448\\

\bottomrule
\end{tabular}
\end{threeparttable}
\end{table}

\section{More Results}
\label{more_results}
\textbf{Full Results on Benchmark.}
\label{full_results_on_benchmark}
We compare the performance of \workname~to baselines using different base LLMs.
Since our approach requires model fine-tuning, we implement it only on LLMs with fewer than 7B parameters. However, we also provide the performance of baselines using 70B parameter LLMs and other advanced commercial LLMs as a reference.
Tab.~\ref{tab:main_results_benchmark_all} shows the full experiment results on \benchmark.
Results show that \workname~consistently outperforms the baselines when using the same LLMs.
We also observe that the baseline approach ICL-All using GPT-4o can achieve the highest performance with 92.1\% Acc-P, 71.5\% F1-score, and 59.6\% Acc-Args.
Additionally, \workname~with a 7B parameter LLM achieves performance comparable to the best baseline using a 70B LLM, with only 0.4\% lower Acc-P, 1.5\% lower F1-score, and 0.4\% higher Acc-Args, respectively, demonstrating the strong performance of \workname.

\begin{table}[t]
\setlength\tabcolsep{5pt}
\centering
\scriptsize
\caption{
Main results on \benchmarklite.
}
\label{tab:main_results_benchmark_lite}
\begin{threeparttable}
\begin{tabular}{c|c|c|c|c|c|c|c|c|c}
\toprule
\multirow{2}{*}{\textbf{Model}}     & \multirow{2}{*}{\textbf{Method}} &\multicolumn{4}{c|}{\textbf{Proactive Predictions}}  & \multicolumn{4}{c}{\textbf{Tool Calling}}  \\ \cmidrule(l){3-10} 
&  & \textbf{Acc-P$^{\uparrow}$} &\textbf{MD.$^{\downarrow}$}
&\textbf{FD.$^{\downarrow}$}  & \textbf{RMSE$^{\downarrow}$}   & \textbf{Precision$^{\uparrow}$}  & \textbf{Recall$^{\uparrow}$}   & \textbf{F1-score$^{\uparrow}$}  & \textbf{Acc-Args$^{\uparrow}$}  \\ 
\midrule
\multirow{5}{*}{GPT-4o}    & Proactive Agent& 0.720& 0.260& 0.020 & 2.481& 0.630 & 0.531 & 0.555& 0.235\\
& Vanilla ICL   & 0.710&0.270 &0.020  &2.517 & 0.600 & 0.525 & 0.544&0.153\\
& CoT  & 0.680& 0.310& 0.010 &2.630 & 0.607 & 0.557 & 0.571&0.207\\
& ICL-P  & 0.850& 0.130& 0.020 &1.770 & 0.755 & 0.635 & 0.667&0.352\\
& ICL-All  & 0.890& 0.110& 0.000 &1.627 & 0.782 & 0.675 & 0.701& 0.397\\

\midrule
\multirow{5}{*}{GPT-3.5-Turbo}    & Proactive Agent& 0.510& 0.010& 0.480 & 2.174& 0.220 & 0.128 & 0.152& 0.114\\
& Vanilla ICL   & 0.630& 0.050& 0.320 & 2.253& 0.410 & 0.282 & 0.314&0.172\\
& CoT  & 0.660& 0.190& 0.150 &2.186 & 0.531 & 0.439 & 0.464&0.126\\
& ICL-P  & 0.710& 0.020& 0.270 & 1.854& 0.545 & 0.415 &0.450 &0.213\\
& ICL-All  & 0.830& 0.060& 0.110 & 1.578& 0.635 & 0.537 &0.561 &0.216 \\

\midrule
\multirow{5}{*}{Qwen2.5-72B-Ins}    & Proactive Agent& 0.780& 0.180& 0.040 & 2.258& 0.670 & 0.550 & 0.582& 0.143\\
& Vanilla ICL   & 0.770& 0.120& 0.110 & 2.000& 0.595 & 0.481 & 0.512& 0.136\\
& CoT  & 0.770& 0.120& 0.110 &1.786 & 0.575 & 0.467 & 0.494& 0.158\\
& ICL-P  & 0.760& 0.030& 0.210 & 1.612& 0.605 & 0.471 & 0.506& 0.205\\
& ICL-All  &0.840 & 0.020& 0.140 & 1.349& 0.646 & 0.530 & 0.559& 0.221\\

\midrule
\multirow{5}{*}{Llama3.1-70B-Ins}   & Proactive Agent& 0.760& 0.150& 0.090 &2.076 & 0.620 & 0.529 & 0.554& 0.253\\
& Vanilla ICL   & 0.740& 0.050& 0.210 & 1.889& 0.530 & 0.421 & 0.449& 0.222\\
& CoT  & 0.800& 0.060& 0.140 &1.649 & 0.585 &0.491  &0.515 & 0.222\\
& ICL-P  & 0.720& 0.010& 0.270 & 1.841& 0.546 & 0.426 & 0.457& 0.255\\
& ICL-All  & 0.857& 0.035& 0.107 & 1.300& 0.684 & 0.581 & 0.605& 0.310\\

\midrule
\multirow{7}{*}{DeepSeek-R1-7B}     & Proactive Agent& 0.540& 0.420& 0.040 & 3.119& 0.490 & 0.470 & 0.475&0.039 \\
& Vanilla ICL   & 0.540& 0.310& 0.150 & 2.849& 0.455 & 0.411 & 0.423& 0.103\\
& CoT  & 0.560& 0.320& 0.120 & 2.796& 0.465 & 0.447 & 0.452& 0.070\\
& ICL-P  & 0.610&  0.240 & 0.150& 2.624 & 0.445& 0.400& 0.413& 0.083\\
& ICL-All  & 0.650& 0.260& 0.090 & 2.541& 0.455 & 0.434 & 0.439& 0.080\\
& Vanilla SFT  & 0.760& 0.120&  0.120& 1.786& 0.581&  0.561& 0.559& 0.406\\
&\cellcolor{mygray}\textit{\workname}  & \cellcolor{mygray}0.840& \cellcolor{mygray}0.050&  \cellcolor{mygray}0.110& \cellcolor{mygray}1.510& \cellcolor{mygray}0.678& \cellcolor{mygray}0.641& \cellcolor{mygray}0.648& \cellcolor{mygray}0.487\\

\midrule
\multirow{7}{*}{Qwen2.5-7B-Ins}     & Proactive Agent& 0.710& 0.210& 0.080 &2.328 & 0.575 & 0.515 & 0.532& 0.064\\
& Vanilla ICL   & 0.700& 0.280& 0.020 & 2.596& 0.595 & 0.533 & 0.550& 0.107\\
& CoT  & 0.750& 0.230& 0.020 &2.306 & 0.630 &0.541  &0.564 & 0.163\\
& ICL-P  & 0.840& 0.080& 0.080 & 1.783& 0.656 & 0.570 & 0.595& 0.224\\
& ICL-All  & 0.850& 0.100& 0.050 & 1.780& 0.615 & 0.565 & 0.573& 0.183\\
& Vanilla SFT & 0.640& 0.190& 0.170 & 2.206& 0.520 & 0.457 & 0.476& 
0.342 \\
& \cellcolor{mygray}\textit{\workname}  &\cellcolor{mygray}0.850 &\cellcolor{mygray}0.050& \cellcolor{mygray}0.100& \cellcolor{mygray}1.403& \cellcolor{mygray}0.667& \cellcolor{mygray}0.615& \cellcolor{mygray}0.624& \cellcolor{mygray}0.479\\

\midrule
\multirow{7}{*}{Llama3.1-8B-Ins}     & Proactive Agent& 0.490& 0.000& 0.510 & 2.469& 0.200 & 0.088 & 0.117&0.082 \\
& Vanilla ICL   & 0.580& 0.320& 0.100 & 2.623& 0.500 & 0.450 & 0.466& 0.136\\
& CoT  & 0.630& 0.360& 0.010 & 2.306& 0.595 & 0.553 & 0.564& 0.166\\
& ICL-P  & 0.710&  0.210 & 0.080& 2.315 & 0.535& 0.495& 0.506& 0.155\\
& ICL-All  & 0.770& 0.170& 0.060 & 1.757& 0.598 & 0.526 & 0.543& 0.177\\
& Vanilla SFT  & 0.720& 0.120& 0.160 & 1.959& 0.536 & 0.497 & 0.508&0.362\\
&\cellcolor{mygray}\textit{\workname}  & \cellcolor{mygray}0.830& \cellcolor{mygray}0.070&  \cellcolor{mygray}0.100& \cellcolor{mygray}1.510& \cellcolor{mygray}0.687& \cellcolor{mygray}0.637& \cellcolor{mygray}0.645& \cellcolor{mygray}0.486\\

\bottomrule
\end{tabular}
\end{threeparttable}
\end{table}

\begin{table}[t]
\setlength\tabcolsep{5.8pt}
\centering
\scriptsize
\caption{
Overall performance of \workname~using different LLMs as base models.
}
\label{tab:baseLLM}
\begin{threeparttable}
\begin{tabular}{@{}c|c|c|c|c|c|c|c|c|c|c@{}}
\toprule
\multirow{2}{*}{\textbf{Category}}     & \multirow{2}{*}{\textbf{Model}} & \multirow{2}{*}{\textbf{Size}} & \multicolumn{4}{c|}{\textbf{Proactive Predictions}}  & \multicolumn{4}{c}{\textbf{Tool Calling}} \\ \cmidrule(l){4-11} 
&  &  & \textbf{Acc-P$^{\uparrow}$}  & \textbf{MD.$^{\downarrow}$}
&\textbf{FD.$^{\downarrow}$}  & \textbf{RMSE$^{\downarrow}$}   & \textbf{Precision$^{\uparrow}$}  & \textbf{Recall$^{\uparrow}$}   & \textbf{F1-score$^{\uparrow}$}  & \textbf{Acc-Args$^{\uparrow}$} \\
\midrule
\multirow{3}{*}{7B $\sim$ 8B}    & LLaMA3  & 8B & 0.898 & 0.028& 0.074& 1.254 & 0.685 & 0.653 & 0.652& 0.465\\
& Qwen2.5   & 7B & 0.883 &0.040 &0.076 &  1.215  &0.682 & 0.653 & 0.653 & 0.481 \\
& DeepSeek-R1   & 7B & 0.888 & 0.038&0.074  & 1.275& 0.659 & 0.648 &0.639 &0.434 \\
\midrule
\multirow{3}{*}{1.5B $\sim$ 3B} & Qwen2.5     & 3B& 0.869& 0.043& 0.086 &  1.336  & 0.652& 0.610 & 0.615 & 0.421 \\
& DeepSeek-R1  & 1.5B    & 0.882& 0.041& 0.076 &  1.245  &0.686 & 0.652 &0.652 & 0.447 \\
& Qwen2.5  & 1.5B    &0.869 &0.038 & 0.091&  1.312  &0.642 &0.612  & 0.613 &0.410  \\

\bottomrule
\end{tabular}
\end{threeparttable}
\end{table}

\begin{table}[t]
\setlength\tabcolsep{5pt}
\centering
\scriptsize
\caption{
Results on out-of-domain experiments.
}
\label{tab:main_results_ood}
\begin{threeparttable}
\begin{tabular}{c|c|c|c|c|c|c|c|c|c}

\toprule
\multirow{2}{*}{\textbf{Model}}     & \multirow{2}{*}{\textbf{Method}} &\multicolumn{4}{c|}{\textbf{Proactive Predictions}}  & \multicolumn{4}{c}{\textbf{Tool Calling}}  \\ \cmidrule(l){3-10} 
&  & \textbf{Acc-P$^{\uparrow}$} &\textbf{MD.$^{\downarrow}$}
&\textbf{FD.$^{\downarrow}$}  & \textbf{RMSE$^{\downarrow}$}   & \textbf{Precision$^{\uparrow}$}  & \textbf{Recall$^{\uparrow}$}   & \textbf{F1-score$^{\uparrow}$}  & \textbf{Acc-Args$^{\uparrow}$}  \\ 
\midrule
\multirow{5}{*}{GPT-4o}    & Proactive Agent& 0.749& 0.234& 0.016 & 2.446& 0.652 & 0.542 & 0.568& 0.331\\
& Vanilla ICL   & 0.779&0.207 &0.013  &2.253 & 0.681 & 0.573 & 0.600&0.338\\
& CoT  & 0.756& 0.234& 0.010 &2.407 & 0.659 & 0.565 & 0.587&0.306\\
& ICL-P  & 0.909& 0.076& 0.013 &1.514 & 0.735 & 0.619 & 0.648&0.393\\
& ICL-All  & 0.886& 0.100& 0.013 &1.701 & 0.719 & 0.611 & 0.639& 0.397\\

\midrule
\multirow{5}{*}{GPT-3.5-Turbo}    & Proactive Agent& 0.682& 0.010& 0.307 & 1.757& 0.478 & 0.352 & 0.384& 0.126\\
& Vanilla ICL   & 0.742& 0.050& 0.207 & 1.797& 0.555 & 0.412 & 0.448&0.222\\
& CoT  & 0.802& 0.183& 0.013 &2.152 & 0.658 & 0.549 & 0.576&0.206\\
& ICL-P  & 0.702& 0.000& 0.297 & 1.589& 0.521 & 0.376 &0.412 &0.207\\
& ICL-All  & 0.879& 0.020& 0.100 & 1.452& 0.657 & 0.521 &0.555 &0.235 \\

\midrule
\multirow{5}{*}{Qwen2.5-72B-Ins}    & Proactive Agent& 0.846& 0.070& 0.083 & 1.806& 0.657 & 0.545 & 0.573& 0.240\\
& Vanilla ICL   & 0.792& 0.090& 0.117 & 1.600& 0.627 & 0.522 & 0.551& 0.330\\
& CoT  & 0.819& 0.140& 0.040 &1.564 & 0.678 & 0.583 & 0.607& 0.330\\
& ICL-P  & 0.882& 0.013& 0.103 & 1.199& 0.675 & 0.569 & 0.595& 0.297\\
& ICL-All  &0.929 & 0.020& 0.050 & 1.036& 0.717 & 0.618 & 0.642& 0.355\\

\midrule
\multirow{5}{*}{Llama3.1-70B-Ins}   & Proactive Agent& 0.796& 0.107& 0.097 & 1.950& 0.659 & 0.551 & 0.580&0.292 \\
& Vanilla ICL   &0.689 & 0.020& 0.291 & 1.840& 0.445 & 0.350 & 0.377& 0.221\\
& CoT  & 0.829& 0.026& 0.143 & 1.386& 0.629 & 0.489 & 0.525& 0.089\\
& ICL-P  & 0.725& 0.003& 0.271 & 1.734&  0.533& 0.423 & 0.453& 0.184\\
& ICL-All  & 0.926& 0.003& 0.070 & 1.127& 0.756 & 0.626 & 0.660& 0.327\\

\midrule
\multirow{7}{*}{DeepSeek-R1-7B}     & Proactive Agent& 0.571& 0.391& 0.036 & 3.040& 0.475 & 0.453 & 0.459&0.018 \\
& Vanilla ICL   & 0.665& 0.254& 0.080 & 2.621& 0.483 & 0.435 & 0.447& 0.085\\
& CoT  & 0.605& 0.347& 0.046 & 2.777& 0.516 & 0.467 & 0.479& 0.076\\
& ICL-P  & 0.675&  0.237 & 0.087& 2.527 & 0.489& 0.438& 0.450& 0.069\\
& ICL-All  & 0.639& 0.321& 0.040 & 2.637& 0.505 & 0.466 & 0.476& 0.071\\
& Vanilla SFT  & 0.782& 0.087&  0.130& 1.743& 0.585&  0.565& 0.564& 0.418\\
&\cellcolor{mygray}\textit{\workname}  & \cellcolor{mygray}0.893& \cellcolor{mygray}0.026&  \cellcolor{mygray}0.080& \cellcolor{mygray}1.249& \cellcolor{mygray}0.681& \cellcolor{mygray}0.645& \cellcolor{mygray}0.648& \cellcolor{mygray}0.489\\

\midrule
\multirow{7}{*}{Qwen2.5-7B-Ins}     & Proactive Agent& 0.806& 0.147& 0.046 &2.139 & 0.605 & 0.556 & 0.567& 0.050\\
& Vanilla ICL   & 0.829& 0.120& 0.050 & 1.898& 0.597 & 0.532 & 0.546& 0.207\\
& CoT  & 0.846& 0.123& 0.030 &1.836 & 0.612 &0.545  &0.562 & 0.218\\
& ICL-P  & 0.886& 0.050& 0.063 & 1.584& 0.569 & 0.517 & 0.530& 0.192\\
& ICL-All  & 0.890& 0.040& 0.070 & 1.502& 0.593 & 0.527 & 0.543& 0.206\\
& Vanilla SFT  & 0.799& 0.077& 0.123 & 1.685& 0.607 & 0.585 & 0.582& 
0.433 \\
& \cellcolor{mygray}\textit{\workname}  &\cellcolor{mygray}0.909 &\cellcolor{mygray}0.020& \cellcolor{mygray}0.070& \cellcolor{mygray}1.172& \cellcolor{mygray}0.711& \cellcolor{mygray}0.699& \cellcolor{mygray}0.689& \cellcolor{mygray}0.516\\

\midrule
\multirow{7}{*}{Llama3.1-8B-Ins}     & Proactive Agent& 0.652& 0.013& 0.334 & 1.918& 0.403 & 0.266 & 0.301&0.139 \\
& Vanilla ICL   & 0.739& 0.163& 0.097 & 2.052& 0.565 & 0.470 & 0.496& 0.210\\
& CoT  & 0.732& 0.220& 0.046 & 1.934& 0.623 & 0.529 & 0.554& 0.247\\
& ICL-P  & 0.839&  0.137 & 0.023& 1.907 & 0.659& 0.558& 0.585& 0.320\\
& ICL-All  & 0.796& 0.173& 0.030 & 1.822& 0.623 & 0.527 & 0.553& 0.246\\
& Vanilla SFT  & 0.816& 0.090& 0.093 & 1.161& 0.628 & 0.619 & 0.611&0.460\\
&\cellcolor{mygray}\textit{\workname}  & \cellcolor{mygray}0.869& \cellcolor{mygray}0.036&  \cellcolor{mygray}0.093& \cellcolor{mygray}1.369& \cellcolor{mygray}0.661& \cellcolor{mygray}0.645& \cellcolor{mygray}0.641& \cellcolor{mygray}0.480\\

\bottomrule
\end{tabular}
\end{threeparttable}
\end{table}

Tab.~\ref{tab:main_results_benchmark_lite} shows the full results on \benchmarklite.
Tab.~\ref{tab:main_results_ood} shows the full results on \benchmark~under OOD settings.
Similarly, we observe that \workname~consistently achieves the highest performance when using the same base LLMs as the baselines.
In addition, \workname~can still achieve performance comparable to the baselines that use 70B scale LLMs.
For example, \workname~using Qwen2.5-7B-Ins can achieve 0.7\% lower Acc-P, 1.9\% higher F1-score, and 16.9\% higher Acc-Args compared to the best baseline with Llama3.1-70B-Ins.
Results validate the effectiveness of \workname.

\textbf{Impact of Thought Traces on ICL.}
We also analyze the effectiveness of thought traces on ICL.
First, we observe that CoT generally 
outperforms Vanilla ICL on \benchmark, both in in-domain and OOD settings.
This indicates that incorporating thought traces into few-shot demonstrations can enhance the performance of ICL.
Additionally, we observe that ICL-All also outperforms ICL-P most of the time.
This suggests that even after integrating persona context, further incorporating thought traces can offer benefits to ICL approaches for this task.
Furthermore, we observe that thought traces provide greater benefits for 70B-sized LLMs compared to 7B LLMs. For instance, using Llama3.1-70B-Ins, ICL-All achieves a 20.1\% higher Acc-P, a 20.7\% higher F1-score, and a 14.3\% higher Acc-Args than ICL-P on \benchmark~under OOD settings.
This may be because the limited parameters in smaller LLMs result in inherently limited knowledge, making it challenging for them to fully learn the distilled thought traces from more advanced LLMs.

\textbf{Impact of Different Base Models.}
We test using different LLMs as the base model in \workname.
Table~\ref{tab:baseLLM} shows that 1.5B to 3B LLMs (e.g., Qwen2.5‑1.5B‑Instruct) perform only 2.9\%, 3.9\%, and 5.5\% lower on Acc‑P, F1‑score, and Acc‑Args, respectively, compared to 7B LLMs.
This reveals the opportunities to deploy \workname~on mobile devices without accessing the cloud, further reducing privacy concerns and system overhead~\cite{yang2023edgefm}.

Tab.~\ref{tab:ablation_study_benchmark} shows the ablation study using different base LLMs. The results show that persona context is crucial for the task. Removing it from \workname~can lead to significant decreases across all metrics, with Acc-P and Acc-Args reductions of up to 12.0\% and 14.3\%, respectively.
Additionally, sensory context perception and thought traces can also bring positive benefits.

\begin{figure}
  \centering
\includegraphics[width=1\linewidth]{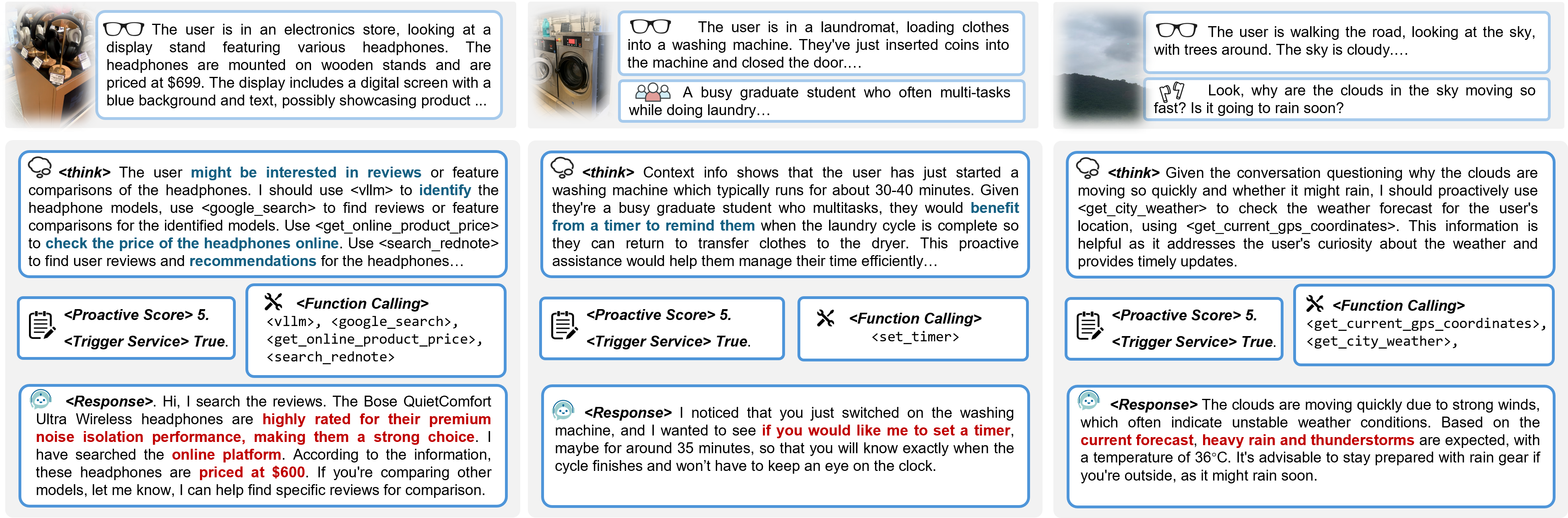}
\caption{Qualitative results of \workname~in proactive cases, showing sensory and persona contexts, and \workname's thought traces, proactive predictions, tool calls, and final responses.}  \label{fig:qualitative_results_overall_proactive_supp}
\end{figure}

\begin{figure}
  \centering
\includegraphics[width=1\linewidth]{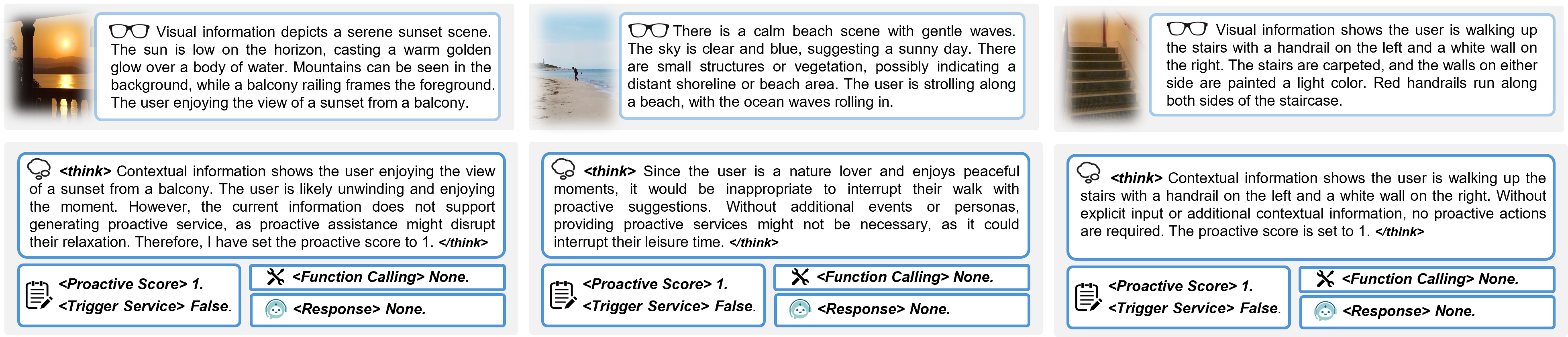}
\vspace{-15pt}
\caption{Qualitative results of \workname~in non-proactive cases.}  \label{fig:qualitative_results_overall_noproactive_supp}
\vspace{-10pt}
\end{figure}

\begin{table}[t]
\setlength\tabcolsep{3.6pt}
\centering
\scriptsize
\caption{
Ablation study on \benchmark.
``w/o persona'' means the agent does not use persona information during both the training and testing stages. 
``w/o think'' means that the SFT training data does not include the thought process.
}
\label{tab:ablation_study_benchmark}
\begin{threeparttable}
\begin{tabular}{c|c|c|c|c|c|c|c|c|c}
\toprule
\multirow{2}{*}{\textbf{Model}}     & \multirow{2}{*}{\textbf{Method}} &\multicolumn{4}{c|}{\textbf{Proactive Predictions}}  & \multicolumn{4}{c}{\textbf{Tool Calling}}  \\ \cmidrule(l){3-10} 
&  & \textbf{Acc-P$^{\uparrow}$} &\textbf{MD.$^{\downarrow}$}
&\textbf{FD.$^{\downarrow}$}  & \textbf{RMSE$^{\downarrow}$}   & \textbf{Precision$^{\uparrow}$}  & \textbf{Recall$^{\uparrow}$}   & \textbf{F1-score$^{\uparrow}$}  & \textbf{Acc-Args$^{\uparrow}$}  \\ 

\midrule
\multirow{3}{*}{Qwen2.5-7B} & w/o persona           & 0.775& 0.078& 0.146 & 1.799 & 0.571& 0.531 &0.532 &0.364\\
&w/o think   &0.857 & 0.030& 0.112 & 1.415 & 0.634& 0.592 & 0.599& 0.411\\
& \cellcolor{mygray}\workname  &\cellcolor{mygray}0.894 &\cellcolor{mygray}0.013 & \cellcolor{mygray}0.091 & \cellcolor{mygray}1.264& \cellcolor{mygray}0.672 & \cellcolor{mygray}0.644 & \cellcolor{mygray}0.645& \cellcolor{mygray}0.459\\

\midrule
\multirow{3}{*}{Llama3.1-8B}    & w/o persona   & 0.806 & 0.081& 0.112 & 1.639 & 0.611& 0.574 &0.579 &0.405\\
& w/o think  & 0.857 & 0.030& 0.112 & 1.397 & 0.645& 0.607 & 0.612& 0.419\\
& \cellcolor{mygray}\workname  & \cellcolor{mygray}0.874& \cellcolor{mygray}0.030&  \cellcolor{mygray}0.095& \cellcolor{mygray}1.408& \cellcolor{mygray}0.660& \cellcolor{mygray}0.627& \cellcolor{mygray}0.626& \cellcolor{mygray}0.448\\

\midrule
\multirow{3}{*}{DeepSeek-R1-7B}   & w/o persona  & 0.799& 0.092& 0.109 & 1.742 & 0.609& 0.575 & 0.580&0.409\\
& w/o think  & 0.884& 0.017& 0.099 & 1.355 & 0.652&0.625  &0.625 & 0.439\\
& \cellcolor{mygray}\workname  & \cellcolor{mygray}0.888& \cellcolor{mygray}0.027&  \cellcolor{mygray}0.085& \cellcolor{mygray}1.319& \cellcolor{mygray}0.676& \cellcolor{mygray}0.648& \cellcolor{mygray}0.647& \cellcolor{mygray}0.468\\

\bottomrule
\end{tabular}
\end{threeparttable}
\end{table}

\textbf{Performance Across Different Tool Chain Lengths.}
Although Tab.~\ref{tab:main_results_benchmark_all} shows that \workname~outperforms the baselines on \benchmark, we provide a detailed analysis in this section.
Specifically, we examine the performance of \workname~across samples with varying tool chain lengths, categorizing them into three groups: 0–1 tools (level 1), 2 tools (level 2), and 3–5 tools (level 3), as shown in 
Tab.~\ref{tab:main_results_benchmark_level1}, Tab.~\ref{tab:main_results_benchmark_level2}, and Tab.~\ref{tab:main_results_benchmark_level3}, respectively.
Results show that \workname~achieves significantly higher F1-score, and Acc-Args than the baselines on the level 2 and level 3 samples.
For level 3 samples, \workname~can achieve 30.3\% and 16.0\% higher Acc-P and Acc-Args, respectively, when using Qwen2.5-7B-Ins as the base LLM, validating its strong performance in scenarios that require multiple external tools for proactive services.

\begin{table}[t]
\setlength\tabcolsep{5pt}
\centering
\scriptsize
\caption{
Results for the Level-1 samples in \benchmark.
}
\label{tab:main_results_benchmark_level1}
\begin{threeparttable}
\begin{tabular}{c|c|c|c|c|c|c|c|c|c}
\toprule
\multirow{2}{*}{\textbf{Model}}     & \multirow{2}{*}{\textbf{Method}} &\multicolumn{4}{c|}{\textbf{Proactive Predictions}}  & \multicolumn{4}{c}{\textbf{Tool Calling}}  \\ \cmidrule(l){3-10} 
&  & \textbf{Acc-P$^{\uparrow}$} &\textbf{MD.$^{\downarrow}$}
&\textbf{FD.$^{\downarrow}$}  & \textbf{RMSE$^{\downarrow}$}   & \textbf{Precision$^{\uparrow}$}  & \textbf{Recall$^{\uparrow}$}   & \textbf{F1-score$^{\uparrow}$}  & \textbf{Acc-Args$^{\uparrow}$}  \\ 

\midrule
\multirow{5}{*}{GPT-4o}    & Proactive Agent& 0.835& 0.138& 0.025 &1.983 & 0.789 & 0.789 & 0.789& 0.246\\
& Vanilla ICL & 0.835& 0.138& 0.025 & 1.997& 0.776 & 0.784 &0.778 & 0.462\\
& CoT  & 0.835& 0.128& 0.035 &1.970 & 0.775 &0.784  &0.777 &0.529 \\
& ICL-P  & 0.928& 0.046& 0.025 &1.432 & 0.860 & 0.866 &0.862 &0.596 \\
& ICL-All  & 0.923& 0.046& 0.030 & 1.421& 0.839 & 0.851 & 0.842& 0.596\\

\midrule
\multirow{5}{*}{GPT-3.5-Turbo}    & Proactive Agent& 0.517& 0.005& 0.476 & 2.104& 0.467 & 0.467 & 0.467& 0.065 \\
& Vanilla ICL   & 0.702& 0.051& 0.246 & 1.820& 0.617 & 0.620 & 0.617& 0.374\\
& CoT  & 0.815 &0.071 & 0.112 & 1.525& 0.706 & 0.717 & 0.710& 0.433\\
& ICL-P  & 0.733& 0.035& 0.230 &1.872 & 0.625 & 0.630 &0.627 &0.400 \\
& ICL-All  & 0.800& 0.051& 0.148 & 1.542& 0.709 & 0.717 & 0.712& 0.458\\

\midrule
\multirow{5}{*}{Qwen2.5-72B-Ins}    & Proactive Agent& 0.800& 0.061& 0.069 & 1.684& 0.700 & 0.702 & 0.700& 0.144\\
& Vanilla ICL   & 0.835& 0.082& 0.082 & 1.353& 0.729 & 0.743 & 0.734& 0.475\\
& CoT  & 0.856& 0.076& 0.066 & 1.254& 0.747 & 0.759 & 0.751& 0.468\\
& ICL-P  & 0.887& 0.046& 0.066 &1.081 & 0.805 & 0.810 & 0.806& 0.495\\
& ICL-All  & 0.923& 0.035& 0.041 & 0.981& 0.806 & 0.815 & 0.808& 0.474\\

\midrule
\multirow{5}{*}{Llama3.1-70B-Ins}   & Proactive Agent& 0.825& 0.041& 0.133 & 1.676& 0.687 & 0.692 & 0.688& 0.314\\
& Vanilla ICL   & 0.697& 0.020& 0.282 & 1.688& 0.548 & 0.553 & 0.550& 0.374\\
& CoT  & 0.774& 0.030& 0.194 & 1.394& 0.612 & 0.625 &0.616 &0.385 \\
& ICL-P  & 0.784& 0.000& 0.215 & 1.529& 0.654 & 0.666 & 0.658& 0.374\\
& ICL-All  & 0.861& 0.005& 0.133 & 1.176& 0.736 & 0.748 & 0.740& 0.455 \\

\midrule
\multirow{7}{*}{DeepSeek-R1-7B}     & Proactive Agent& 0.707& 0.225& 0.066 & 2.444& 0.671 & 0.671 & 0.671&0.019 \\
& Vanilla ICL   & 0.697& 0.143& 0.159 & 2.215& 0.588 & 0.594 & 0.590& 0.078\\
& CoT  & 0.759& 0.200& 0.041 & 2.328& 0.705 & 0.707 & 0.706& 0.109\\
& ICL-P  & 0.759&  0.117 & 0.123& 2.065 & 0.658& 0.666& 0.660& 0.126\\
& ICL-All  & 0.825& 0.133& 0.041 & 1.978& 0.714 & 0.717 & 0.715& 0.103\\
& Vanilla SFT  & 0.779& 0.056&  0.164& 1.783& 0.628&  0.651& 0.635& 0.386\\
&\cellcolor{mygray}\textit{\workname}  & \cellcolor{mygray}0.856& \cellcolor{mygray}0.015&  \cellcolor{mygray}0.128& \cellcolor{mygray}1.423& \cellcolor{mygray}0.712& \cellcolor{mygray}0.748& \cellcolor{mygray}0.723& \cellcolor{mygray}0.468\\

\midrule
\multirow{7}{*}{Qwen2.5-7B-Ins}     & Proactive Agent& 0.820& 0.082& 0.097 & 1.843& 0.692 & 0.702 & 0.695& 0.038\\
& Vanilla ICL   & 0.769& 0.087& 0.143 & 1.868& 0.642 & 0.661 & 0.648& 0.326\\
& CoT  & 0.794& 0.082& 0.123 & 1.874& 0.668 & 0.676 & 0.670& 0.272\\
& ICL-P  & 0.805& 0.082& 0.112 & 1.845& 0.683 & 0.697 & 0.688& 0.303\\
& ICL-All  &0.851 & 0.082& 0.066 & 1.724& 0.735 & 0.748 & 0.739& 0.301\\
& Vanilla SFT  & 0.733& 0.061& 0.205 & 1.905& 0.588 & 0.605 & 0.593&0.398\\
& \cellcolor{mygray}\textit{\workname}  &\cellcolor{mygray}0.856 &\cellcolor{mygray}0.005 & \cellcolor{mygray}0.138 & \cellcolor{mygray}1.396& \cellcolor{mygray}0.701 & \cellcolor{mygray}0.728 & \cellcolor{mygray}0.709& \cellcolor{mygray}0.459\\

\midrule
\multirow{7}{*}{Llama3.1-8B-Ins}     & Proactive Agent& 0.533& 0.010& 0.456 & 2.169 & 0.328 &0.328 &0.328 & 0.091 \\
& Vanilla ICL   & 0.794& 0.117& 0.087 & 1.593& 0.707 & 0.707 & 0.707& 0.234\\
& CoT  & 0.774& 0.117& 0.107 & 1.637& 0.659 &0.661  & 0.659& 0.257\\
& ICL-P  & 0.794& 0.194& 0.010 &1.900 &0.753  & 0.753 & 0.753& 0.251\\
& ICL-All  & 0.861& 0.097& 0.041 &1.379 & 0.748 & 0.748 & 0.748& 0.298\\
& Vanilla SFT  & 0.774& 0.061& 0.164 & 1.778& 0.635 & 0.671 & 0.647&0.362\\
&\cellcolor{mygray}\textit{\workname}  & \cellcolor{mygray}0.841& \cellcolor{mygray}0.015&  \cellcolor{mygray}0.143& \cellcolor{mygray}1.485& \cellcolor{mygray}0.695& \cellcolor{mygray}0.733& \cellcolor{mygray}0.707& \cellcolor{mygray}0.448\\

\bottomrule
\end{tabular}
\end{threeparttable}
\end{table}

\begin{table}[t]
\setlength\tabcolsep{5pt}
\centering
\scriptsize
\caption{
Results for the Level-2 samples in \benchmark.
}
\label{tab:main_results_benchmark_level2}
\begin{threeparttable}
\begin{tabular}{c|c|c|c|c|c|c|c|c|c}
\toprule
\multirow{2}{*}{\textbf{Model}}     & \multirow{2}{*}{\textbf{Method}} &\multicolumn{4}{c|}{\textbf{Proactive Predictions}}  & \multicolumn{4}{c}{\textbf{Tool Calling}}  \\ \cmidrule(l){3-10} 
&  & \textbf{Acc-P$^{\uparrow}$} &\textbf{MD.$^{\downarrow}$}
&\textbf{FD.$^{\downarrow}$}  & \textbf{RMSE$^{\downarrow}$}   & \textbf{Precision$^{\uparrow}$}  & \textbf{Recall$^{\uparrow}$}   & \textbf{F1-score$^{\uparrow}$}  & \textbf{Acc-Args$^{\uparrow}$}  \\ 

\midrule
\multirow{5}{*}{GPT-4o}    & Proactive Agent& 0.651& 0.348& 0.000 &2.761 & 0.500 & 0.255 & 0.337& 0.256\\
& Vanilla ICL & 0.674& 0.325& 0.000 & 2.676& 0.333 & 0.209 &0.249 & 0.467\\
& CoT  & 0.720& 0.279& 0.000 &2.645 & 0.472 &0.325  &0.370 &0.534 \\
& ICL-P  & 0.883& 0.116& 0.000 &1.855 & 0.511 & 0.290 &0.364 &0.606 \\
& ICL-All  & 0.883& 0.116& 0.000 & 1.848& 0.546 & 0.314 & 0.391& 0.590\\

\midrule
\multirow{5}{*}{GPT-3.5-Turbo}    & Proactive Agent& 0.953& 0.046& 0.000 & 1.229& 0.581 & 0.290 & 0.387& 0.070 \\
& Vanilla ICL   & 0.814& 0.186& 0.000 & 1.867& 0.407 & 0.220 & 0.282& 0.366\\
& CoT  & 0.790 &0.209 & 0.000 & 1.994& 0.480 & 0.302 & 0.359& 0.449\\
& ICL-P  & 0.883& 0.116& 0.000 &1.830 & 0.534 & 0.279 &0.364 &0.408 \\
& ICL-All  & 0.883& 0.116& 0.000 & 1.532& 0.492 & 0.325 & 0.379& 0.472\\

\midrule
\multirow{5}{*}{Qwen2.5-72B-Ins}    & Proactive Agent& 0.930& 0.069& 0.000 & 1.758& 0.534 & 0.302 & 0.379& 0.150\\
& Vanilla ICL   & 0.883& 0.116& 0.000 & 1.509& 0.484 & 0.372 & 0.405& 0.484\\
& CoT  & 0.907& 0.093& 0.000 & 1.430& 0.511 & 0.372 & 0.418& 0.478\\
& ICL-P  & 0.930& 0.069& 0.000 &1.355 & 0.639 & 0.476 & 0.527& 0.506\\
& ICL-All  & 0.907& 0.093& 0.000 & 1.486& 0.542 & 0.453 & 0.479& 0.493\\

\midrule
\multirow{5}{*}{Llama3.1-70B-Ins}   & Proactive Agent& 0.837& 0.162& 0.000 & 1.935& 0.503 & 0.302 & 0.367& 0.333\\
& Vanilla ICL   & 1.000& 0.000& 0.000 & 0.849& 0.531 & 0.395 & 0.439& 0.381\\
& CoT  & 1.000& 0.000& 0.000 & 0.902& 0.589 & 0.348 &0.427 &0.392 \\
& ICL-P  & 1.000& 0.000& 0.000 & 0.821& 0.624 & 0.418 & 0.486& 0.385\\
& ICL-All  & 1.000& 0.000& 0.000 & 0.876& 0.662 & 0.476 & 0.532& 0.465 \\

\midrule
\multirow{7}{*}{DeepSeek-R1-7B}     & Proactive Agent& 0.279& 0.720& 0.000 & 3.882& 0.058 & 0.034 & 0.042&0.021 \\
& Vanilla ICL   & 0.465& 0.534& 0.000 & 3.348& 0.129 & 0.104 & 0.110& 0.082\\
& CoT  & 0.348& 0.651& 0.000 & 3.592& 0.104 & 0.081 & 0.089& 0.115\\
& ICL-P  & 0.465&  0.534 & 0.000& 3.306 & 0.089& 0.058& 0.067& 0.129\\
& ICL-All  & 0.418& 0.581& 0.000 & 3.474& 0.127 & 0.081 & 0.096& 0.096\\
& Vanilla SFT  & 0.860& 0.139&  0.000& 1.486& 0.472&  0.337& 0.377& 0.404\\
&\cellcolor{mygray}\textit{\workname}  & \cellcolor{mygray}0.930& \cellcolor{mygray}0.069&  \cellcolor{mygray}0.000& \cellcolor{mygray}1.181& \cellcolor{mygray}0.536& \cellcolor{mygray}0.441& \cellcolor{mygray}0.464& \cellcolor{mygray}0.471\\

\midrule
\multirow{7}{*}{Qwen2.5-7B-Ins}     & Proactive Agent& 0.674& 0.325& 0.000 & 2.663& 0.302 & 0.220 & 0.248& 0.040\\
& Vanilla ICL   & 0.907& 0.093& 0.000 & 1.372& 0.461 & 0.383 & 0.408& 0.324\\
& CoT  & 0.930& 0.069& 0.000 & 1.422& 0.418 & 0.279 & 0.325& 0.270\\
& ICL-P  & 0.860& 0.139& 0.000 & 1.861& 0.434 & 0.314 & 0.351& 0.302\\
& ICL-All  &0.883 & 0.116& 0.000 & 1.758& 0.476 & 0.314 & 0.368& 0.316\\
& Vanilla SFT  & 0.744& 0.255& 0.000 & 1.823& 0.412 & 0.372 & 0.377&0.419\\
& \cellcolor{mygray}\textit{\workname}  &\cellcolor{mygray}0.977 &\cellcolor{mygray}0.023 & \cellcolor{mygray}0.000 & \cellcolor{mygray}0.927& \cellcolor{mygray}0.542 & \cellcolor{mygray}0.465 & \cellcolor{mygray}0.479& \cellcolor{mygray}0.479\\

\midrule
\multirow{7}{*}{Llama3.1-8B-Ins}     & Proactive Agent& 1.000& 0.000& 0.000 & 0.940 & 0.383 &0.197 &0.259 & 0.088 \\
& Vanilla ICL   & 0.558& 0.441& 0.000 & 2.583& 0.290 & 0.162 & 0.205& 0.231\\
& CoT  & 0.674& 0.325& 0.000 & 2.151& 0.352 &0.209  & 0.255& 0.252\\
& ICL-P  & 0.534& 0.465& 0.000 &2.672 &0.418  & 0.232 & 0.294& 0.241\\
& ICL-All  & 0.651& 0.348& 0.000 &2.327 & 0.383 & 0.232 & 0.282& 0.291\\
& Vanilla SFT  & 0.860& 0.139& 0.000 & 1.599& 0.395 & 0.325 & 0.348&0.371\\
&\cellcolor{mygray}\textit{\workname}  & \cellcolor{mygray}0.907& \cellcolor{mygray}0.093&  \cellcolor{mygray}0.000& \cellcolor{mygray}1.462& \cellcolor{mygray}0.503& \cellcolor{mygray}0.430& \cellcolor{mygray}0.445& \cellcolor{mygray}0.468\\

\bottomrule
\end{tabular}
\end{threeparttable}
\end{table}

\begin{table}[t]
\setlength\tabcolsep{5pt}
\centering
\scriptsize
\caption{
Results for the Level-3 samples in \benchmark.
}
\label{tab:main_results_benchmark_level3}
\begin{threeparttable}
\begin{tabular}{c|c|c|c|c|c|c|c|c|c}
\toprule
\multirow{2}{*}{\textbf{Model}}     & \multirow{2}{*}{\textbf{Method}} &\multicolumn{4}{c|}{\textbf{Proactive Predictions}}  & \multicolumn{4}{c}{\textbf{Tool Calling}}  \\ \cmidrule(l){3-10} 
&  & \textbf{Acc-P$^{\uparrow}$} &\textbf{MD.$^{\downarrow}$}
&\textbf{FD.$^{\downarrow}$}  & \textbf{RMSE$^{\downarrow}$}   & \textbf{Precision$^{\uparrow}$}  & \textbf{Recall$^{\uparrow}$}   & \textbf{F1-score$^{\uparrow}$}  & \textbf{Acc-Args$^{\uparrow}$}  \\ 

\midrule
\multirow{5}{*}{GPT-4o}    & Proactive Agent& 0.714& 0.285& 0.000 &2.604 & 0.523 & 0.158 & 0.239& 0.253\\
& Vanilla ICL & 0.857& 0.142& 0.000 & 1.817& 0.732 & 0.332 &0.434 & 0.466\\
& CoT  & 0.910& 0.089& 0.000 &1.564 & 0.784 &0.406  &0.504 &0.534 \\
& ICL-P  & 0.928& 0.071& 0.000 &1.463 & 0.794 & 0.357 &0.467 &0.598 \\
& ICL-All  & 0.946& 0.053& 0.000 & 1.309& 0.794 & 0.415 & 0.516& 0.598\\

\midrule
\multirow{5}{*}{GPT-3.5-Turbo}    & Proactive Agent& 1.000& 0.000& 0.000 & 0.845& 0.517 & 0.149 & 0.231& 0.068 \\
& Vanilla ICL   & 0.982& 0.017& 0.000 & 1.069& 0.702 & 0.244 & 0.351& 0.377\\
& CoT  & 0.946 &0.053 & 0.000 & 1.093& 0.647 & 0.337 & 0.422& 0.440\\
& ICL-P  & 1.000& 0.000& 0.000 &0.790 & 0.741 & 0.305 &0.408 &0.408 \\
& ICL-All  & 0.981& 0.017& 0.000 & 0.886& 0.727 & 0.374 & 0.462& 0.468\\

\midrule
\multirow{5}{*}{Qwen2.5-72B-Ins}    & Proactive Agent& 0.928& 0.071& 0.000 & 1.797& 0.669 & 0.237 & 0.343& 0.144\\
& Vanilla ICL   & 0.964& 0.035& 0.000 & 1.043& 0.738 & 0.407 & 0.503& 0.482\\
& CoT  & 0.964& 0.035& 0.000 & 1.101& 0.720 & 0.451 & 0.528& 0.477\\
& ICL-P  & 1.000& 0.000& 0.000 &0.790 & 0.770 & 0.422 & 0.518& 0.504\\
& ICL-All  & 1.000& 0.000& 0.000 & 0.916& 0.717 & 0.428 & 0.511& 0.487\\

\midrule
\multirow{5}{*}{Llama3.1-70B-Ins}   & Proactive Agent& 0.946& 0.053& 0.000 & 1.232& 0.595 & 0.268 & 0.356& 0.318\\
& Vanilla ICL   & 1.000& 0.000& 0.000 & 0.668& 0.738 & 0.398 & 0.497& 0.381\\
& CoT  & 1.000& 0.000& 0.000 & 0.801& 0.758 & 0.407 &0.497 &0.392 \\
& ICL-P  & 1.000& 0.000& 0.000 & 0.719& 0.708 & 0.394 & 0.480& 0.386\\
& ICL-All  & 1.000& 0.000& 0.000 & 0.707& 0.668 & 0.410 & 0.482& 0.460 \\

\midrule
\multirow{7}{*}{DeepSeek-R1-7B}     & Proactive Agent& 0.285& 0.714& 0.000 & 3.964& 0.142 & 0.048 & 0.070&0.040 \\
& Vanilla ICL   & 0.607& 0.392& 0.000 & 2.991& 0.251 & 0.125 & 0.158& 0.081\\
& CoT  & 0.517& 0.482& 0.000 & 3.348& 0.238 & 0.105 & 0.137& 0.111\\
& ICL-P  & 0.625&  0.375 & 0.000& 2.945 & 0.363& 0.148& 0.203& 0.130\\
& ICL-All  & 0.500& 0.500& 0.000 & 3.313& 0.279 & 0.117 & 0.158& 0.098\\
& Vanilla SFT  & 0.946& 0.053&  0.000& 1.093& 0.714&  0.467& 0.542& 0.400\\
&\cellcolor{mygray}\textit{\workname}  & \cellcolor{mygray}1.000& \cellcolor{mygray}0.000&  \cellcolor{mygray}0.000& \cellcolor{mygray}0.755& \cellcolor{mygray}0.721& \cellcolor{mygray}0.444& \cellcolor{mygray}0.529& \cellcolor{mygray}0.455\\

\midrule
\multirow{7}{*}{Qwen2.5-7B-Ins}     & Proactive Agent& 0.821& 0.178& 0.000 & 2.129& 0.392 & 0.199 & 0.255& 0.039\\
& Vanilla ICL   & 0.910& 0.089& 0.000 & 1.586& 0.506 & 0.263 & 0.334& 0.332\\
& CoT  & 0.892& 0.107& 0.000 & 1.742& 0.425 & 0.196 & 0.258& 0.271\\
& ICL-P  & 0.910& 0.089& 0.000 & 1.690& 0.488 & 0.251 & 0.316& 0.306\\
& ICL-All  &0.910 & 0.089& 0.000 & 1.679& 0.410 & 0.184 & 0.246& 0.305\\
& Vanilla SFT  & 0.946& 0.053& 0.000 & 1.157& 0.731 & 0.504 & 0.576&0.417\\
& \cellcolor{mygray}\textit{\workname}  &\cellcolor{mygray}0.964 &\cellcolor{mygray}0.035 & \cellcolor{mygray}0.000 & \cellcolor{mygray}0.972& \cellcolor{mygray}0.672 & \cellcolor{mygray}0.489 & \cellcolor{mygray}0.549& \cellcolor{mygray}0.465\\

\midrule
\multirow{7}{*}{Llama3.1-8B-Ins}     & Proactive Agent& 1.000& 0.000& 0.000 & 0.731 & 0.538 &0.177 &0.260 & 0.090 \\
& Vanilla ICL   & 0.732& 0.267& 0.000 & 2.228& 0.526 & 0.174 & 0.257& 0.235\\
& CoT  & 0.821& 0.178& 0.000 & 1.889& 0.567 &0.225  & 0.308& 0.264\\
& ICL-P  & 0.625& 0.375& 0.000 &2.741 &0.508  & 0.153 & 0.233& 0.245\\
& ICL-All  & 0.875& 0.125& 0.000 &1.752 & 0.574 & 0.222 & 0.306& 0.299\\
& Vanilla SFT  & 0.946& 0.053& 0.000 & 1.149& 0.620 & 0.389 & 0.459&0.369\\
&\cellcolor{mygray}\textit{\workname}  & \cellcolor{mygray}0.964& \cellcolor{mygray}0.035&  \cellcolor{mygray}0.000& \cellcolor{mygray}1.043& \cellcolor{mygray}0.657& \cellcolor{mygray}0.406& \cellcolor{mygray}0.477& \cellcolor{mygray}0.466\\

\bottomrule
\end{tabular}
\end{threeparttable}
\end{table}

\textbf{More Qualitative Results.}
We also provide more qualitative results of \workname~in
Fig.~\ref{fig:qualitative_results_overall_proactive_supp} and Fig.~\ref{fig:qualitative_results_overall_noproactive_supp}.
Results indicate that \workname~initiates the proactive support when it perceives contexts such as the user showing interest in a product or putting clothes into a washing machine.
Additionally, integrating knowledge from external tools, such as online product prices and app-based recommendations, can further enhance the usefulness of these proactive services.
Moreover, when \workname~detects contexts like leisure time enjoying the sunset or beach waves, or normal daily activities such as walking upstairs, it correctly identifies that no proactive support is needed, thus reducing unnecessary interruptions.

\section{Limitations}
\label{limitations}
Although \workname~demonstrates strong performance in the context-aware proactive agent task, we summarize the limitations and future directions for this work as follows.
First, the current tool set in \workname~includes twenty tools with APIs. The recent emergence of the Model Context Protocol (MCP) aims to standardize these external tools, which can be integrated into \workname~to further enhance the standardization of tool calls and the diversity of tool types.
Second, although \benchmark~currently includes nine life scenarios, it can be further diversified to enhance the \workname's capabilities and practicality for daily use.

\section{Broader Impacts}
\label{broader_impacts}
In this paper, we explore the context-aware proactive LLM agents for the first time, and propose a framework called \workname.
By utilizing the rich contextual information from sensory perceptions alongside tool-based LLM reasoning, \workname~significantly enhances both perception and functionality compared to existing approaches, resulting in improved proactive service.
\workname~utilizes sensor data from wearable devices such as smart glasses and earphones. This hands-free, egocentric perception not only offers a better understanding of the user's conditions and intentions but also reduces both cognitive and physical workload, perfectly aligning with the vision of a proactive assistant.
In addition, to bridge the gap in evaluating this new task, we introduce \benchmark, the first benchmark to evaluate context-aware proactive LLM agents.
Furthermore, \workname~serves as a bridge between research on sensory context perception in ubiquitous mobile systems and the emerging LLM agents, thereby opening up new research perspectives and directions.
We hope our research will help advance the development of proactive, human-centric AI assistants.

\end{document}